\documentclass[letterpaper]{article}
\usepackage[utf8]{inputenc}

\usepackage{aaai2027} 
\nocopyright
\usepackage[hyphens]{url}  
\usepackage{graphicx} 
\usepackage{natbib}  
\usepackage{caption}
\usepackage{algorithm}
\usepackage{algorithmic}
\usepackage{amssymb}
\usepackage{newfloat}
\usepackage{amsmath}
\usepackage{listings}
\usepackage{multirow}
\usepackage{amsfonts}
\usepackage{amssymb}

\usepackage{booktabs}
\usepackage{multirow}
\usepackage{array}

\DeclareCaptionStyle{ruled}{labelfont=normalfont,labelsep=colon,strut=off} 
\floatstyle{ruled}
\newfloat{listing}{tb}{lst}{}
\floatname{listing}{Listing}

\usepackage{booktabs}

\title{VideoVIBE: A Video-Grounded Diagnostic Benchmark for One-Shot Interactive Website Generation}

\author{
Jiajun Xu\textsuperscript{\rm 1},
Yanghao Zhou\textsuperscript{\rm 2}$^{\ddagger}$,
Jingyun Liao\textsuperscript{\rm 3},
Yu Bai\textsuperscript{\rm 6},
Jinxing Zhou\textsuperscript{\rm 4},\\
Chengliang Liu\textsuperscript{\rm 7},
Changsen Yuan\textsuperscript{\rm 5},
Bo Wang\textsuperscript{\rm 2},
Qian Liu\textsuperscript{\rm 8}
}

\affiliations{
\textsuperscript{\rm 1}University of Technology Sydney,
\textsuperscript{\rm 2}Beijing Institute of Technology,
\textsuperscript{\rm 3}Hunan University,
\textsuperscript{\rm 4}OpenNLP Lab\\
\textsuperscript{\rm 5}Beijing University of Technology,
\textsuperscript{\rm 6}Beijing Academy of Artificial Intelligence,
\textsuperscript{\rm 7}University of Macau,
\textsuperscript{\rm 8}University of Auckland\\
$^{\ddagger}$Project Leader.
}

\newcommand{\scorecell}[1]{\makebox[2.65em][c]{#1}}

\begin{document}

\maketitle
\begin{abstract}
Natural-language-driven ``vibe coding'' enables the one-shot generation of visually rich and interactive web applications, yet reliable assessment of their quality has not kept pace. Existing evaluations often score isolated artifacts or final task outcomes, offering limited evidence about which failures occur and why. We introduce \textbf{VideoVIBE}, a video-grounded benchmark that transforms human-operated webpage recordings into fine-grained diagnostic tasks. It contains approximately 1.7K diagnostic Video QA instances derived from 6,338 verified failures across generated webpages, spanning semantic-logical, visual-motion, structural-temporal, and functional failures. Diagnoses are grounded primarily in recorded presentation and behavior, with webpage source code used as complementary context. We further propose \textbf{V2Lens}, a training-free, evidence-grounded multi-agent system that challenges and selectively refines initial video-based diagnoses through targeted visual and source-code verification. Across thirteen closed-source and open-weight Video MLLMs, Gemini-2.5-Flash is the strongest standalone model with a score of 64.54, while V2Lens reaches 71.72, an improvement of 7.18 points. Together, our results show that video-grounded evaluation can move beyond isolated artifacts and aggregate outcomes toward a behaviorally faithful and diagnostically informative account of generated application quality.
\end{abstract}

\begin{figure}[t]
    \centering

    \includegraphics[width=\columnwidth]{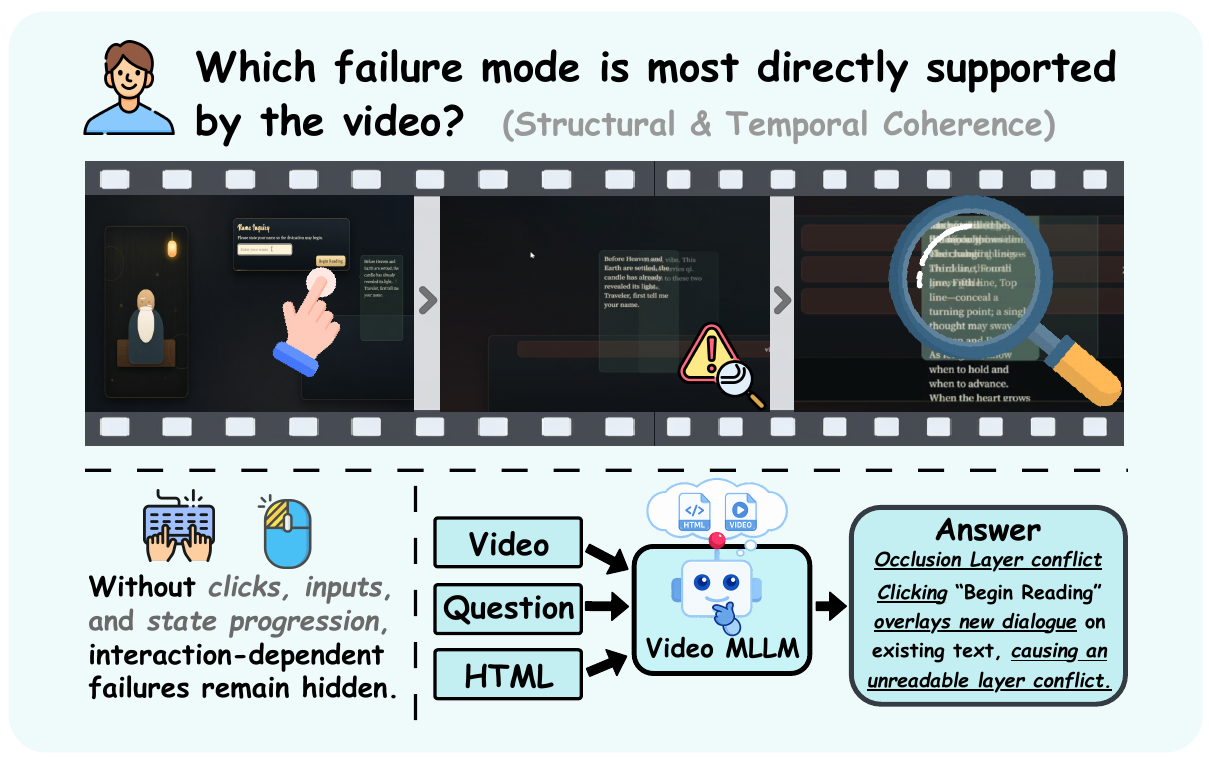}
    \caption{Overview of the VideoVIBE diagnostic task. In this example, a layer conflict becomes visible only after the user enters information and clicks ``Begin Reading'', causing a new dialog to overlap existing content. Given the interaction video and diagnostic question, together with complementary webpage source HTML code, a Video MLLM selects the failure mode best supported by the evidence.}
    \label{fig:intro}
\end{figure}

\section{Introduction}

Coding agents and natural-language-driven ``vibe coding'' are reshaping software development by enabling users to translate high-level intent directly into executable applications~\cite{lu2026webgen,liu2026webcoderbench,li2026mm,yang2024swe,wang2025openhands}. Interactive web applications are a particularly important setting because they combine immediate deployability with rich requirements for content, visual design, interaction logic, and application state. In the \textit{one-shot} setting considered in this work, a complete application is generated from a single specification without iterative correction. This capability lowers the barrier to software creation and accelerates rapid prototyping. However, surface completeness does not guarantee behavioral reliability: a visually polished application may still contain inconsistent business rules, broken controls, incorrect state transitions, missing interaction feedback, or failures that emerge only after several user actions. As coding agents increasingly participate in application development, identifying such failures becomes essential for determining whether generated software is dependable in actual use.

Evaluating the quality of these applications remains challenging. Existing protocols commonly inspect static artifacts, including screenshots, rendered interfaces, and source code, or summarize application quality through automated tests, workflow success, and coarse-grained human or model judgments~\cite{yun2024web2code,lin2025webuibench,xiao2025designbench,liu2026webcoderbench,li2025webdevjudge,zhao2026longwebbench,hong2026diageval}. Although useful, these protocols often assess individual artifacts or final outcomes in isolation, providing limited support for identifying concrete failure cases and distinguishing their underlying modes. Content inconsistencies, poor legibility, and disrupted visual hierarchy may be obscured by aggregate quality scores, while erroneous state transitions, navigation mappings, and broken controls may become apparent only during execution. A comprehensive evaluation should therefore diagnose failures across both rendered interface states and application behavior over time.

Human-operated interaction recordings provide a common source of evidence for these complementary aspects. Individual frames preserve webpage content, layout, typography, visual hierarchy, and motion feedback, while the complete sequence captures user inputs, interface responses, state transitions, navigation behavior, and functional execution~\cite{videowebarena,lin2024videogui,dai2026webvr,xu2026video2code}. Fig.~\ref{fig:intro} illustrates an interaction-dependent example that cannot be identified from the initial rendered state alone. After the user enters the required information and clicks ``Begin Reading'', the webpage transitions to a new state in which the newly opened dialog overlaps existing content, making the text unreadable. Other failures, such as inconsistent content, poor contrast, or inappropriate visual hierarchy, can instead be diagnosed directly from rendered frames. Together, frame-level and trajectory-level evidence enable a unified protocol for diagnosing both static presentation defects and interaction-dependent failures.

Based on this formulation, we introduce \textbf{VideoVIBE}, a video-grounded diagnostic benchmark for one-shot-generated interactive
web applications. VideoVIBE evaluates webpage failures as manifested in recorded application sessions rather than judging source code as a
standalone artifact. To ensure diversity and controlled comparison, we apply a shared set of user-goal-oriented prompts across seven
representative web-generation systems and record human interactions with the resulting applications. From the retained recordings, we identify 6,338 manually verified failure events and organize them into a hierarchical taxonomy comprising four top-level dimensions, twelve evaluation subdimensions, and thirty fine-grained failure modes. The four dimensions cover \emph{semantic and logical consistency}, \emph{visual and motion fidelity}, \emph{structural and temporal coherence}, and \emph{functional executability}. Representative selection further yields approximately 1.7K diagnostic Video QA instances.
Each diagnostic instance asks a model to select the fine-grained failure mode best supported by the evidence. The interaction video provides primary evidence of rendered appearance and observed behavior, while the source code offers complementary implementation-level context for distinguishing similar manifestations with different underlying mechanisms; it is not independently scored. VideoVIBE therefore evaluates evidence-grounded failure-mode discrimination rather than generic source-code inspection, binary failure detection, or end-to-end task success alone.

To improve diagnostic reliability, we propose \textbf{V2Lens}, a training-free, evidence-grounded multi-agent system. V2Lens derives an initial video diagnosis, then uses targeted source-code inspection and tool-assisted visual analysis to evaluate alternative hypotheses. Its gated refinement process replaces the initial prediction only when a challenge answer is sufficiently supported and independently confirmed by a reviewer, integrating visual, temporal, and implementation-level evidence while limiting unsupported revisions.

Experiments across thirteen closed-source and open-weight video multimodal models reveal substantial room for improvement. Among standalone models, the closed-source Gemini-2.5-Flash achieves the highest overall score of 64.54, while Qwen3.5-35B-A3B is the strongest open-weight model with a score of 63.90. No standalone model consistently dominates across all twelve subdimensions, and performance varies across the four failure dimensions, highlighting the difficulty of comprehensive webpage failure diagnosis. Using Gemini-2.5-Flash as its base evaluator, V2Lens improves the overall score from 64.54 to 71.72, a gain of 7.18 points. 
In summary, our contributions are threefold:
\begin{itemize}\item We introduce \textbf{VideoVIBE}, a video-grounded benchmark for diagnosing both static presentation defects and interaction-dependent failures in one-shot-generated web applications. 

\item We develop an expert-guided taxonomy comprising four top-level dimensions, twelve evaluation subdimensions, and thirty fine-grained failure modes, together with a human-verification protocol for constructing evidence-grounded diagnostic questions.

\item We propose \textbf{V2Lens}, a training-free multi-agent system that integrates video analysis, source-code verification, tool-assisted evidence collection, and gated refinement, improving the overall diagnostic score by 7.18 points over its standalone base evaluator.

\end{itemize}

\begin{figure*}[t]
\centering
\includegraphics[width=1\textwidth]{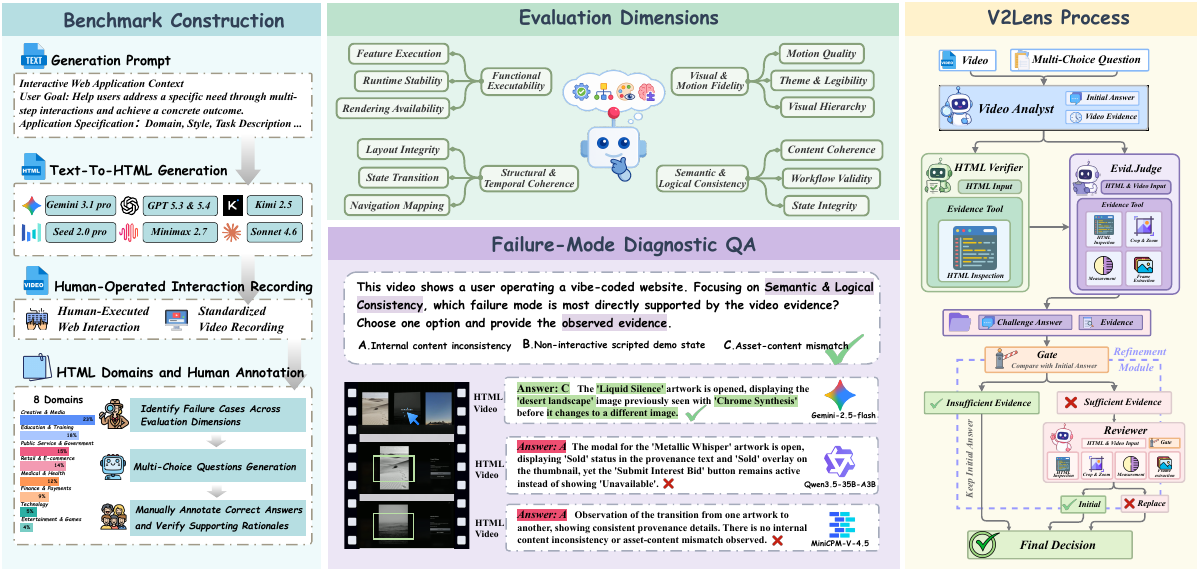} 
\vspace{-1ex}
\vspace{-2ex}
\caption{Overview of VideoVIBE and V2Lens. Left: benchmark construction from prompt design and multi-system webpage generation to human-operated recording, failure annotation, and manually verified diagnostic Video QA construction. Middle: the four evaluation dimensions, twelve subdimensions, and representative failure-mode diagnostic instances grounded in interaction videos and webpage source code. Right: the V2Lens evidence-grounded multi-agent process, which combines video analysis, source-code verification, evidence judging, and gated independent review to produce the final diagnosis.}
\label{fig2}
\vspace{-2ex}
\end{figure*}

\section{Related Work}
\textbf{Text-to-Web Generation.} 
Driven by rapid advances in large language models (LLMs) and multimodal LLMs (MLLMs), web generation has evolved from static webpage understanding and UI-to-code synthesis to interactive, multi-page, and project-level web application construction~\citep{yun2024web2code,lin2025webuibench,xiao2025designbench,awal2025webmmu}. Recent benchmarks evaluate generated applications with complementary protocols: WebGen-Bench executes curated functional test cases using a web-navigation agent~\cite{lu2026webgen},, Vision2Web combines workflow-driven GUI testing with VLM-based visual judging~\cite{vision2web}, and WebCoderBench aggregates rule-based and LLM-based metrics across multiple quality dimensions~\citep{liu2026webcoderbench}. On the generation side, MM-WebAgent employs hierarchical multimodal planning and iterative reflection, whereas WebGen-R1 uses scaffold-driven reinforcement learning with multimodal rewards to improve project-level website generation~\citep{li2026mm,jiang2026webgen}.
These works assess specification satisfaction, functional correctness, visual quality, or aggregate development quality. Their output is typically a task-level score or verdict, rather than a diagnosis of the concrete runtime failure exposed by an interaction trace. In contrast, {VideoVIBE} takes a human-operated interaction recording and the corresponding HTML implementation as evidence, and evaluates whether a Video MLLM can identify the fine-grained failure mode of an already generated application. This formulation complements end-to-end generation evaluation by making dynamic failure diagnosis the target task.

\noindent\textbf{GUI Evaluation and Video-Grounded Web Modeling.}
A line of work evaluates interactive systems through agent trajectories and automated judges. AgentRewardBench~\cite{lu2025agentrewardbench} studies whether LLM judges can assess web-agent trajectories, while DiagEval~\cite{hong2026diageval} uses trajectory-conditioned probes to distinguish evaluator errors from genuine software defects. WebDevJudge~\cite{li2025webdevjudge} evaluates LLMs and MLLMs as static and dynamic web-development critics, while computer-use agents
have been explored as judges of generative user interfaces~\cite{lin2025computer}.
VideoWebArena~\cite{videowebarena} evaluates whether agents can use video tutorials to complete web tasks, whereas WebVR~\cite{dai2026webvr} conditions webpage recreation on screen recordings and assesses reconstruction fidelity with human-aligned visual rubrics. Our {VideoVIBE} differs from these settings in both its evidence and its output. It does not use video as an instruction for task execution or as a target for webpage reconstruction. Instead, it treats a recording as behavioral evidence of an already generated website and requires models to distinguish confusable failure modes across semantic and logical consistency, visual and motion fidelity, structural and temporal coherence, and functional executability. Its target is therefore neither task success, evaluator reliability, nor reconstruction quality, but the specific failure diagnosis supported by the observed interaction.

\section{VideoVIBE Benchmark}

VideoVIBE is a hierarchical diagnostic benchmark for one-shot-generated interactive web applications. Unlike static evaluation based on screenshots or source code alone, it specifically targets failures that emerge as users execute actions, trigger state changes, and complete multi-step workflows. Each benchmark instance combines a human-operated interaction recording with the corresponding webpage source code and requires a model to select the diagnosis best supported by the observed interaction and implementation evidence. Fig.~\ref{fig2} provides an overview of VideoVIBE construction, its diagnostic task, and the V2Lens process.

\subsection{Task Definition}
Let $\mathcal{D}_i = \left(V_i, S_i, q_i, \mathcal{A}_i, y_i\right)$ denote the $i$-th diagnostic instance, where $V_i$ is a human-operated interaction video, $S_i$ is the corresponding webpage source code, represented as a self-contained HTML document, $q_i$ is a diagnostic question, $\mathcal{A}_i$ is a set of candidate failure modes, and $y_i \in \mathcal{A}_i$ is the ground-truth diagnosis. At evaluation time, a model receives $\left(V_i, S_i, q_i, \mathcal{A}_i\right)$ and predicts $\hat{y}_i \in \mathcal{A}_i$.

\begin{figure*}[!t]
    \centering

    \includegraphics[width=\textwidth]
    {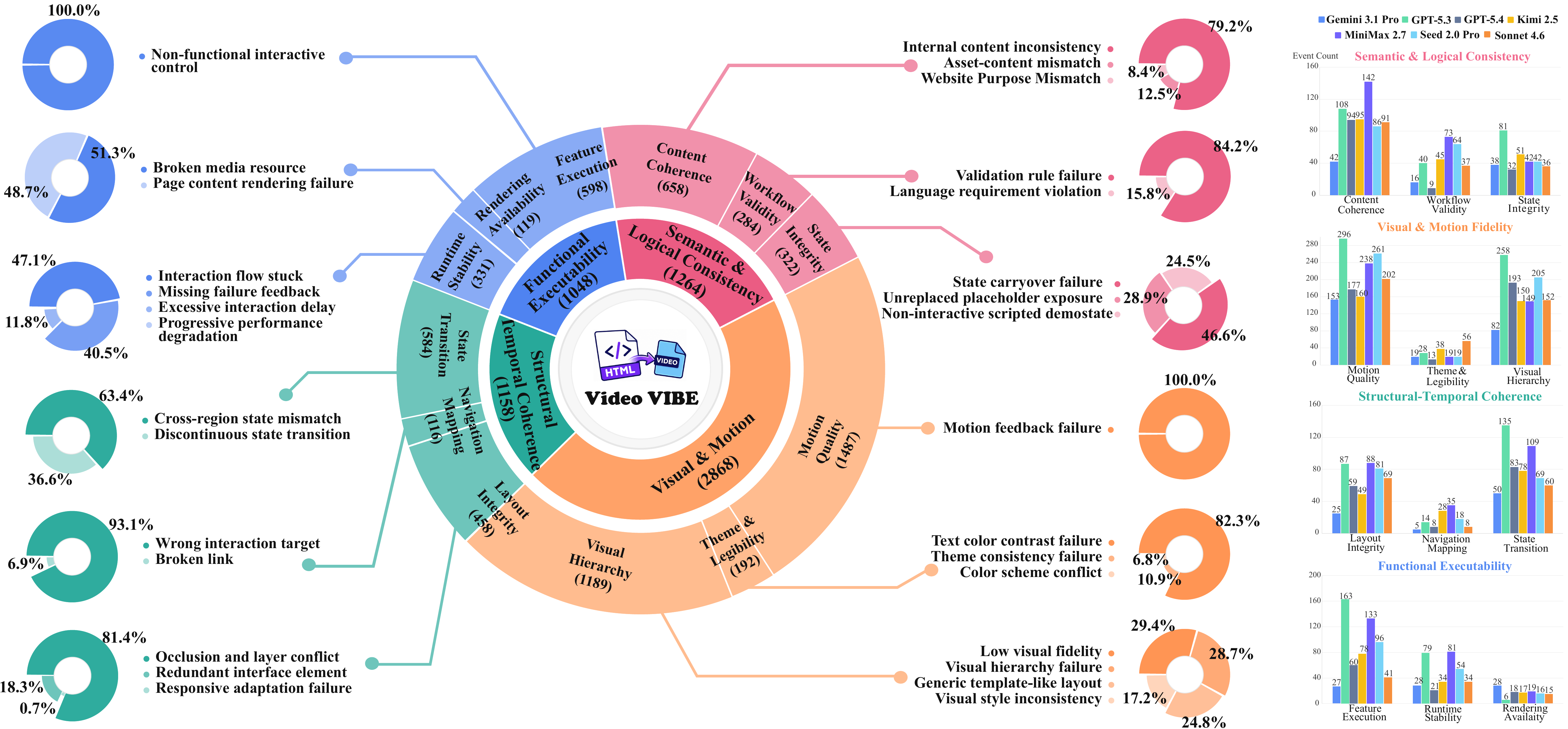}

    \par\smallskip

    \begin{minipage}[t]{0.82\textwidth}
        \centering
        {\small
        (a) Hierarchical failure taxonomy and case distribution.
        }
    \end{minipage}
    \hfill
    \begin{minipage}[t]{0.17\textwidth}
        \centering
        {\small
        (b) Failure distribution.
        }
    \end{minipage}
    \caption{Failure taxonomy and distribution of VideoVIBE. (a) Three-level taxonomy comprising four top-level dimensions, twelve evaluation subdimensions, and thirty fine-grained failure modes. Numbers in parentheses denote verified failure-event counts, while the outer donut charts show the distribution of fine-grained modes within each subdimension. (b) Comparison of verified failure-event counts across twelve evaluation subdimensions for seven webpage-generation systems.}
    \label{fig:failuremode}
\end{figure*}

The two input modalities provide complementary evidence: $V_i$ captures the user action, interface response, and state evolution, whereas $S_i$ provides implementation-level context for distinguishing failures with similar observable manifestations. Each question targets a manually verified failure event within a designated top-level dimension and presents plausible alternative diagnoses. VideoVIBE therefore evaluates fine-grained failure-mode discrimination rather than binary failure detection or end-to-end task-success evaluation.

\subsection{Failure Taxonomy}

To define the diagnostic label space, we develop an expert-guided and empirically grounded failure taxonomy. Domain experts synthesize recurring error patterns in web generation and inspect representative applications produced in practical LLM-assisted vibe-coding settings. As illustrated in Fig.~\ref{fig:failuremode}(a), the resulting hierarchy comprises four top-level dimensions, twelve evaluation subdimensions, and thirty fine-grained failure modes.

\textbf{Semantic and Logical Consistency} evaluates whether webpage content, business rules, and represented states remain mutually consistent, covering Content Coherence, Workflow Validity, and State Integrity. \textbf{Visual and Motion Fidelity} assesses whether visual presentation and motion feedback faithfully communicate user actions and interface responses through Motion Quality, Theme \& Legibility, and Visual Hierarchy. \textbf{Structural and Temporal Coherence} examines whether layouts, action--target mappings, and successive interface states remain correctly organized throughout interaction, including Layout Integrity, State Transition, and Navigation Mapping. Finally, \textbf{Functional Executability} evaluates whether interactive controls, runtime processes, and required resources operate reliably through Feature Execution, Runtime Stability, and Rendering Availability.

Together, these dimensions distinguish what an application communicates, how it presents feedback, how its structure and state evolve, and whether it executes reliably. The thirty fine-grained failure modes provide the diagnostic labels used for question and candidate-answer construction. Fig.~\ref{fig:failuremode} presents the complete hierarchy and its failure-case distribution, while detailed explanations of evaluation criteria are provided in the Appendix.

\subsection{Benchmark Construction and Quality Control}

\begin{figure}[t]
    \centering
    \includegraphics[width=\columnwidth]{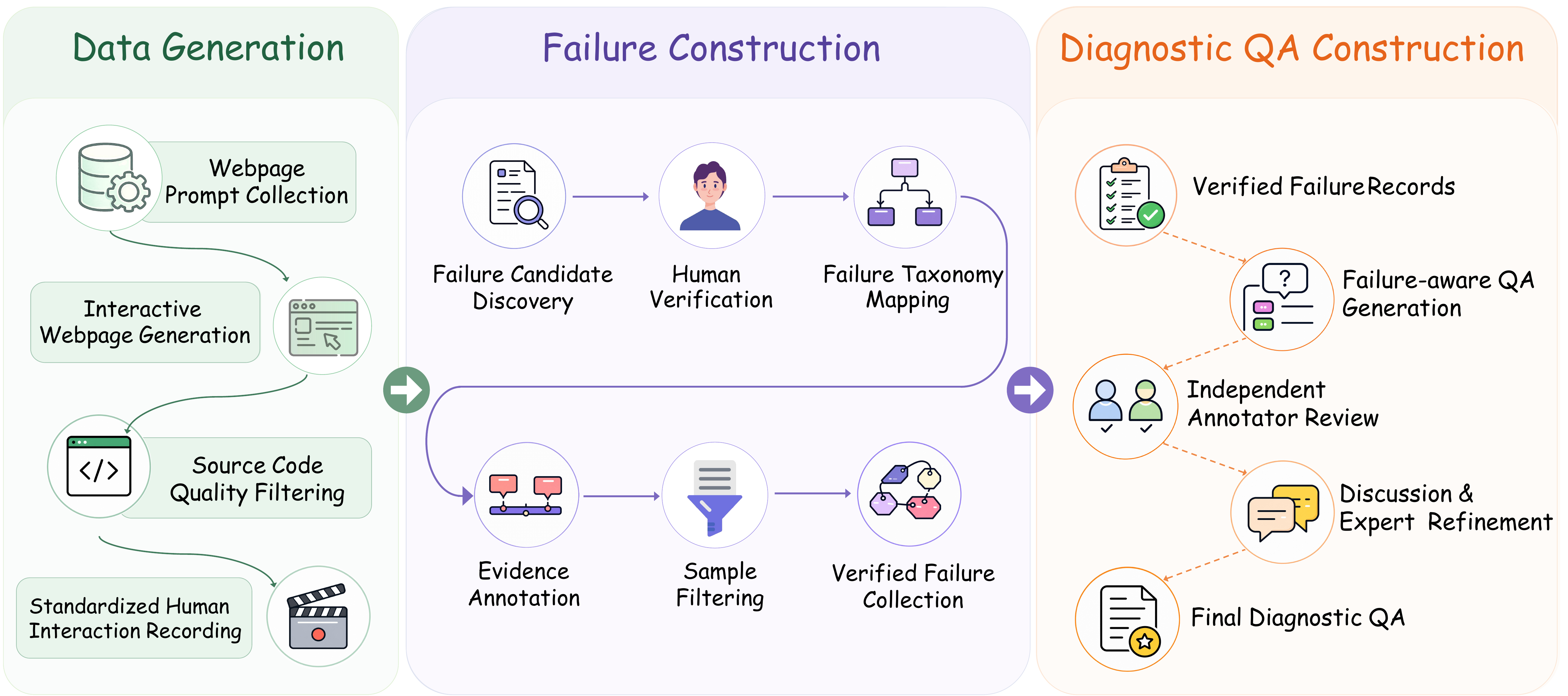}
    \vspace{-2ex}
    \caption{Overview of the VideoVIBE construction and annotation pipeline, from data generation to verified diagnostic QA construction.}
    \label{fig3}
\end{figure}

Fig.~\ref{fig3} summarizes the pipeline from webpage generation to human-verified diagnostic questions.

\noindent\textbf{Webpage Generation and Interaction Recording.}
We construct a shared set of webpage-generation prompts and apply them across seven webpage-generation models to produce diverse interactive applications under controlled conditions. An agent-assisted audit checks whether each application can be reliably initialized, operated, and recorded. Human operators then follow a standardized interaction protocol that exercises the primary functions, state transitions, and multi-step workflows of each application. Recordings containing at least one observable failure are retained for failure annotation and diagnostic QA construction.

\noindent\textbf{Failure Discovery and Annotation.} Automated video analysis identifies candidate anomalies for human review. Annotators then jointly inspect the complete interaction recording and the corresponding webpage source code. A candidate is retained only when its manifestation is observable, reproducible, and unambiguously associated with a fine-grained taxonomy label. For every verified failure event, annotators record its manifestation, supporting evidence, and annotation rationale. Automated tools assist only with candidate discovery, annotation organization, and consistency checking; all final labels and supporting evidence are manually verified.

\noindent\textbf{Diagnostic QA Construction and Verification.} We construct diagnostic question--answer pairs from verified failure events, using LLMs only to assist with the initial drafting of questions and candidate options. To reduce redundancy and prevent videos containing many failures from dominating the benchmark, we treat each video--top-level-dimension pair as the unit of representative selection and retain at most one single-choice question for each pair.
Multiple annotators independently verify the question wording, candidate options, ground-truth diagnosis, and consistency with the supporting evidence. Disagreements are resolved through discussion, while unresolved cases are submitted to expert adjudication. Every retained video is associated with at least one manually verified diagnostic question. Annotation details are provided in the Appendix.

\noindent\textbf{Benchmark Coverage \& Failure Landscape.}
VideoVIBE contains 6,338 manually verified failure events with supporting evidence, spanning all twelve subdimensions and thirty fine-grained failure modes. After representative selection, these events yield approximately 1.7K diagnostic question--answer instances.

Fig.~\ref{fig:failuremode}(b) reports the number of verified failure events for each generator, aggregated from the thirty fine-grained failure modes into twelve evaluation subdimensions. A single video may contain multiple distinct failures, including multiple failures within the same subdimension. Across the seven generators, Motion Quality and Visual Hierarchy exhibit the highest failure counts, followed by Content Coherence, Feature Execution, and State Transition. Their recurrence across multiple systems indicates that these subdimensions represent common challenges in interactive webpage generation rather than model-specific anomalies.

The generators nevertheless exhibit distinct failure profiles. Gemini 3.1 Pro has the fewest failure events, with comparatively fewer failures in Feature Execution, Layout Integrity, and Navigation Mapping. GPT-5.3 and MiniMax 2.7 exhibit higher failure counts but in different areas: GPT-5.3 is more affected by Feature Execution, State Transition, Motion Quality, and Visual Hierarchy, whereas MiniMax 2.7 shows more pronounced weaknesses in Navigation Mapping, Content Coherence, and Workflow Validity. Seed 2.0 Pro exhibits relatively more failures in Motion Quality and Visual Hierarchy, while Claude Sonnet 4.6 shows a more prominent weakness in Theme \& Legibility. These results demonstrate that webpage-generation reliability depends not only on static visual quality, but also on maintaining coherent functionality, interaction feedback, state transitions, and application logic.

\section{V2Lens: Evidence-Grounded Multi-Agent Diagnostic System}
The diverse and model-specific failure profiles revealed by VideoVIBE underscore the challenge of reliably diagnosing one-shot-generated interactive web applications. Such diagnosis requires relating observable interaction behavior to complementary implementation-level evidence and distinguishing among competing failure hypotheses. To this end, we introduce \textbf{V2Lens}, a training-free multi-agent system that performs evidence-grounded verification and conservative answer refinement beyond the one-pass prediction of a standalone multimodal model.
V2Lens treats the interaction video as the primary record of observable behavior and uses the webpage source code as complementary implementation-level evidence. It operates entirely at inference time, requiring neither task-specific training nor parameter updates. As illustrated in Fig.~\ref{fig2} (right panel), V2Lens comprises four specialized agents and two deterministic gates organized into three stages: \textit{initial video diagnosis}, \textit{cross-modal evidence verification}, and \textit{gated refinement}. The system first forms a diagnosis from the complete interaction, then searches for evidence that may challenge it, and revises the prediction only when the proposed alternative passes both evidence-sufficiency screening and independent confirmation.

\begin{table*}[t]
\centering
\caption{
Failure-diagnosis performance (\%) of standalone Video MLLMs and V2Lens on VideoVIBE. Best and second-best results in each column are shown in bold and underlined, respectively. Abbreviations: CC = Content Coherence; WV = Workflow Validity; SI = State Integrity; MQ = Motion Quality; TL = Theme \& Legibility; VH = Visual Hierarchy; LI = Layout Integrity; ST = State Transition; NM = Navigation Mapping; FE = Feature Execution; RS = Runtime Stability; RA = Rendering Availability.
}
\vspace{-2ex}
\label{tab:main_results}

\footnotesize
\setlength{\tabcolsep}{2pt}

\begin{tabular*}{\textwidth}{
@{}
>{\bfseries}l
@{\extracolsep{\fill}}
c
*{12}{c}
@{}
}
\toprule

\textbf{Models}
& \multicolumn{3}{c}{
    {\bfseries\shortstack[c]{Semantic \&\\Logical Consistency}}
}
& \multicolumn{3}{c}{
    {\bfseries\shortstack[c]{Visual \&\\Motion Fidelity}}
}
& \multicolumn{3}{c}{
    {\bfseries\shortstack[c]{Structural \&\\Temporal Coherence}}
}
& \multicolumn{3}{c}{
    {\bfseries\shortstack[c]{Functional\\Executability}}
}
& \scorecell{\textbf{Avg.}}
\\

\cmidrule(lr){2-4}
\cmidrule(lr){5-7}
\cmidrule(lr){8-10}
\cmidrule(lr){11-13}

&
\scorecell{\textbf{CC}}
& \scorecell{\textbf{WV}}
& \scorecell{\textbf{SI}}
& \scorecell{\textbf{MQ}}
& \scorecell{\textbf{TL}}
& \scorecell{\textbf{VH}}
& \scorecell{\textbf{LI}}
& \scorecell{\textbf{ST}}
& \scorecell{\textbf{NM}}
& \scorecell{\textbf{FE}}
& \scorecell{\textbf{RS}}
& \scorecell{\textbf{RA}}
& \scorecell{}
\\
\midrule

\multicolumn{14}{l}{\textbf{Closed-source Models}} \\

Gemini-3.1 Pro
& \scorecell{83.86} & \scorecell{58.70} & \scorecell{\underline{70.54}}
& \scorecell{60.35} & \scorecell{\textbf{67.50}} & \scorecell{39.11}
& \scorecell{77.07} & \scorecell{65.76} & \scorecell{34.15}
& \scorecell{86.10} & \scorecell{68.37} & \scorecell{59.46}
& \scorecell{64.25}
\\

Gemini-3.1-Flash Lite
& \scorecell{76.38} & \scorecell{\underline{72.83}} & \scorecell{60.71}
& \scorecell{59.65} & \scorecell{35.00} & \scorecell{30.67}
& \scorecell{71.97} & \scorecell{\underline{74.46}} & \scorecell{31.71}
& \scorecell{89.84} & \scorecell{60.20} & \scorecell{35.14}
& \scorecell{58.21}
\\

Gemini-3 Flash
& \scorecell{81.50} & \scorecell{66.30} & \scorecell{66.07}
& \scorecell{35.09} & \scorecell{\underline{61.25}} & \scorecell{40.00}
& \scorecell{41.40} & \scorecell{66.30} & \scorecell{14.63}
& \scorecell{83.42} & \scorecell{64.29} & \scorecell{\textbf{64.86}}
& \scorecell{57.09}
\\

Gemini-2.5 Flash
& \scorecell{\underline{85.43}} & \scorecell{69.57} & \scorecell{56.25}
& \scorecell{75.09} & \scorecell{58.75} & \scorecell{43.11}
& \scorecell{68.15} & \scorecell{63.59} & \scorecell{48.78}
& \scorecell{\underline{90.37}} & \scorecell{69.39} & \scorecell{45.95}
& \scorecell{\underline{64.54}}
\\

\midrule
\multicolumn{14}{l}{\textbf{Open-weight models}} \\

VideoLLaMA-3-7B
& \scorecell{46.46} & \scorecell{41.30} & \scorecell{44.64}
& \scorecell{24.21} & \scorecell{18.75} & \scorecell{23.11}
& \scorecell{29.94} & \scorecell{27.72} & \scorecell{24.39}
& \scorecell{52.94} & \scorecell{50.00} & \scorecell{37.84}
& \scorecell{35.11}
\\

MiniCPM-V-4.5-8B
& \scorecell{50.00} & \scorecell{32.61} & \scorecell{58.93}
& \scorecell{34.39} & \scorecell{5.00} & \scorecell{24.89}
& \scorecell{50.96} & \scorecell{53.26} & \scorecell{12.20}
& \scorecell{70.59} & \scorecell{60.20} & \scorecell{37.84}
& \scorecell{40.91}
\\

Qwen3-VL-8B Instruct
& \scorecell{55.51} & \scorecell{48.91} & \scorecell{55.36}
& \scorecell{64.91} & \scorecell{10.00} & \scorecell{39.11}
& \scorecell{50.32} & \scorecell{44.57} & \scorecell{31.71}
& \scorecell{79.68} & \scorecell{43.88} & \scorecell{24.32}
& \scorecell{45.69}
\\

Qwen3-VL-8B Thinking
& \scorecell{74.02} & \scorecell{48.91} & \scorecell{68.75}
& \scorecell{80.00} & \scorecell{28.75} & \scorecell{\textbf{51.11}}
& \scorecell{\underline{78.98}} & \scorecell{57.61} & \scorecell{43.90}
& \scorecell{89.30} & \scorecell{57.14} & \scorecell{40.54}
& \scorecell{59.92}
\\

Qwen3-VL-30B-A3B-Instruct
& \scorecell{61.02} & \scorecell{43.48} & \scorecell{68.75}
& \scorecell{75.79} & \scorecell{13.75} & \scorecell{31.11}
& \scorecell{71.97} & \scorecell{58.15} & \scorecell{26.83}
& \scorecell{86.10} & \scorecell{55.10} & \scorecell{35.14}
& \scorecell{52.27}
\\

Qwen3-VL-30B-A3B-Thinking
& \scorecell{74.80} & \scorecell{47.83} & \scorecell{69.64}
& \scorecell{\underline{84.56}} & \scorecell{27.50} & \scorecell{39.11}
& \scorecell{\textbf{82.80}} & \scorecell{67.93} & \scorecell{46.34}
& \scorecell{88.24} & \scorecell{62.24} & \scorecell{40.54}
& \scorecell{60.96}
\\

Qwen3.5-35B-A3B
& \scorecell{83.46} & \scorecell{66.30} & \scorecell{\textbf{71.43}}
& \scorecell{69.47} & \scorecell{47.50} & \scorecell{36.89}
& \scorecell{63.69} & \scorecell{\textbf{75.54}} & \scorecell{\underline{51.22}}
& \scorecell{83.96} & \scorecell{63.27} & \scorecell{54.05}
& \scorecell{63.90}
\\

GLM-4.6V-106B-A12B
& \scorecell{63.78} & \scorecell{52.17} & \scorecell{58.93}
& \scorecell{63.86} & \scorecell{18.75} & \scorecell{36.89}
& \scorecell{66.88} & \scorecell{58.70} & \scorecell{31.71}
& \scorecell{77.01} & \scorecell{\underline{73.47}} & \scorecell{43.24}
& \scorecell{53.78}
\\

Mimo-v2.5-310B-A15B
& \scorecell{60.63} & \scorecell{42.39} & \scorecell{56.25}
& \scorecell{44.91} & \scorecell{31.25} & \scorecell{36.44}
& \scorecell{61.15} & \scorecell{65.22} & \scorecell{36.59}
& \scorecell{73.26} & \scorecell{52.04} & \scorecell{35.14}
& \scorecell{49.61}
\\

\midrule

\textbf{\texttt{V2Lens (Ours)}}
& \scorecell{\textbf{87.01}} & \scorecell{\textbf{78.26}} & \scorecell{65.18}
& \scorecell{\textbf{85.96}} & \scorecell{58.75} & \scorecell{\underline{47.56}}
& \scorecell{72.61} & \scorecell{\underline{74.46}} & \scorecell{\textbf{58.54}}
& \scorecell{\textbf{93.58}} & \scorecell{\textbf{76.53}} & \scorecell{\underline{62.16}}
& \scorecell{\textbf{71.72}}
\\

\bottomrule
\end{tabular*}
\vspace{-2ex}
\end{table*}

\noindent\textbf{Initial Video Diagnosis.} Given an interaction video and its multiple-choice diagnostic question, the \textbf{\textit{Video Analyst}} examines the complete interaction process and produces an initial answer together with supporting video evidence. It identifies the user actions, interface responses, and state changes relevant to the candidate failure modes, thereby establishing the behavioral evidence against which subsequent source-code findings are evaluated.

\noindent\textbf{Cross-Modal Evidence Verification.} The \textbf{\textit{HTML Verifier}} performs a focused inspection of the webpage source code. It retrieves implementation fragments relevant to the diagnostic question, including document structure, style rules, event handlers, and state-update logic, and assesses whether they support, contradict, or remain inconclusive with respect to each candidate diagnosis. Its findings provide implementation-level context but do not independently establish that a failure manifested during interaction.
The \textbf{\textit{Evidence Judge}} integrates the observations of the Video Analyst, and the findings of the HTML Verifier. When the evidence is insufficient, it may invoke tools for HTML inspection, frame extraction, local-region cropping and enlargement, and visual measurement. These tools enable option-specific examination of both observable behavior and its possible implementation-level causes. The Evidence Judge retains the initial diagnosis unless another candidate is more strongly supported by the collected evidence. When such an alternative exists, it produces a \emph{challenge answer} together with the evidence supporting that challenge.

\noindent\textbf{Gated Refinement.} The challenge answer and its supporting evidence are then passed to the \textit{Refinement Module}. The first deterministic gate, the \textit{Evidence Gate}, compares the challenge answer with the initial answer and applies fixed evidence-sufficiency criteria. A challenge is rejected when it lacks direct behavioral support or relies solely on source-code cues without a corresponding manifestation in the interaction video. In this case, V2Lens retains the initial answer. Challenges supported by sufficient evidence are forwarded to the \textbf{\textit{Reviewer}} for independent verification. The Reviewer operates in a fresh context without access to the Evidence Judge's internal reasoning or rationale. It independently adjudicates between the anonymized initial and challenge answers using the original video, webpage source code, diagnostic question, and candidate options. When necessary, the Reviewer may invoke the same evidence tools to conduct its own inspection. The second deterministic gate, the \textit{Confirmation Gate}, converts the Reviewer's judgment into the final prediction: the challenge answer replaces the initial answer only when it is independently endorsed by the Reviewer; otherwise, the initial answer is retained.
Consequently, no single agent can unilaterally overwrite the initial diagnosis, and V2Lens revises its prediction only when an alternative survives both evidence screening and independent confirmation.

\section{Experiments}
\subsection{Experimental Setup}

\noindent\textbf{Evaluated Models.}
We evaluate thirteen video-capable MLLMs, including four proprietary models and nine open-weight models. The proprietary models comprise  Gemini-3.1-Pro, Gemini-3.1-Flash-Lite, Gemini-3-Flash, and Gemini-2.5-Flash~\cite{comanici2025gemini}. The open-weight models include VideoLLaMA3~\cite{zhang2025videollama}, MiniCPM-V-4.5~\cite{yu2025minicpm}, the Instruct and Thinking variants of Qwen3-VL at the 8B and
30B-A3B scales, and Qwen3.5-35B-A3B~\cite{bai2025qwen3}, GLM-4.6V~\cite{hong2025glm}, and MiMo-V2.5~\cite{li2025xiaomi}. 

\noindent\textbf{Evaluation Protocol.}
We evaluate all models using a unified zero-shot, single-choice diagnostic protocol. Each instance provides an interaction video, the corresponding webpage source code, a diagnostic question, and a set of candidate failure modes. Unless otherwise specified, models jointly analyze the video and source code under the \emph{Video+HTML} setting. All models receive the same task instruction without in-context examples, and their free-form responses are deterministically mapped to the corresponding option labels.

\noindent\textbf{Evaluation Metric.}
Each diagnostic instance belongs to one of the twelve evaluation subdimensions. For subdimension $c$ containing $N_c$ instances, let $y_{c,j}$ and $\hat{y}_{c,j}$ denote the ground-truth and prediction of its $j$-th instance, respectively. We compute the subdimension score and the overall score as
{
\setlength{\abovedisplayskip}{2pt}
\setlength{\belowdisplayskip}{2pt}
\setlength{\jot}{0pt}
\begin{equation}
\begin{aligned}
\mathrm{Score}_c
&=
\frac{1}{N_c}
\sum_{j=1}^{N_c}
\mathbf{1}\!\left[\hat{y}_{c,j}=y_{c,j}\right], \\[-0.4ex]
\mathrm{Overall}
&=
\frac{1}{12}
\sum_{c=1}^{12}
\mathrm{Score}_c.
\end{aligned}
\label{eq:videovibe_metrics}
\end{equation}
Here, $\mathbf{1}[\cdot]$ denotes the indicator function.
We report the twelve
subdimension scores and their unweighted macro-average, giving each
subdimension equal weight regardless of its number of instances.
All scores are reported as percentages. 
}

\begin{table*}[t]
\centering
\caption{
Input-modality ablation for Gemini-2.5-Flash and Qwen3.5-35B-A3B
on VideoVIBE.
}
\label{tab:input_modality_ablation}
\vspace{-1ex}
\footnotesize
\setlength{\tabcolsep}{2.4pt}

\begin{tabular*}{\textwidth}{
@{\extracolsep{\fill}}
l
l
*{12}{c}
c
@{}
}
\toprule

\textbf{Model}
& \textbf{Input}
& \multicolumn{3}{c}{
    \bfseries\shortstack[c]{Semantic \&\\Logical Consistency}
}
& \multicolumn{3}{c}{
    \bfseries\shortstack[c]{Visual \&\\Motion Fidelity}
}
& \multicolumn{3}{c}{
    \bfseries\shortstack[c]{Structural \&\\Temporal Coherence}
}
& \multicolumn{3}{c}{
    \bfseries\shortstack[c]{Functional\\Executability}
}
& \textbf{Avg.}
\\

\cmidrule(lr){3-5}
\cmidrule(lr){6-8}
\cmidrule(lr){9-11}
\cmidrule(lr){12-14}

&
& \textbf{CC}
& \textbf{WV}
& \textbf{SI}
& \textbf{MQ}
& \textbf{TL}
& \textbf{VH}
& \textbf{LI}
& \textbf{ST}
& \textbf{NM}
& \textbf{FE}
& \textbf{RS}
& \textbf{RA}
& \\
\midrule

\multirow{3}{*}{\textbf{Gemini-2.5-Flash}}
& \textbf{Video + HTML}
& \scorecell{\textbf{85.43}}
& \scorecell{\textbf{69.57}}
& \scorecell{\underline{56.25}}
& \scorecell{\textbf{75.09}}
& \scorecell{\underline{58.75}}
& \scorecell{\underline{43.11}}
& \scorecell{\underline{68.15}}
& \scorecell{\underline{63.59}}
& \scorecell{\underline{48.78}}
& \scorecell{\textbf{90.37}}
& \scorecell{\underline{69.39}}
& \scorecell{\underline{45.95}}
& \scorecell{\textbf{64.54}}
\\

& \textbf{Video Only}
& \scorecell{\underline{79.92}}
& \scorecell{50.00}
& \scorecell{\textbf{64.29}}
& \scorecell{\underline{69.47}}
& \scorecell{45.00}
& \scorecell{\textbf{48.44}}
& \scorecell{\textbf{69.43}}
& \scorecell{\textbf{66.85}}
& \scorecell{\textbf{56.10}}
& \scorecell{\underline{88.77}}
& \scorecell{\textbf{77.55}}
& \scorecell{\textbf{54.05}}
& \scorecell{\underline{64.16}}
\\

& \textbf{HTML Only}
& \scorecell{\underline{79.92}}
& \scorecell{\underline{58.70}}
& \scorecell{51.79}
& \scorecell{68.07}
& \scorecell{\textbf{67.50}}
& \scorecell{28.89}
& \scorecell{32.48}
& \scorecell{58.70}
& \scorecell{26.83}
& \scorecell{81.28}
& \scorecell{62.24}
& \scorecell{40.54}
& \scorecell{54.75}
\\

\midrule

\multirow{3}{*}{\textbf{Qwen3.5-35B-A3B}}
& \textbf{Video + HTML}
& \scorecell{\underline{83.46}}
& \scorecell{\textbf{66.30}}
& \scorecell{\textbf{71.43}}
& \scorecell{\textbf{69.47}}
& \scorecell{\underline{47.50}}
& \scorecell{\textbf{36.89}}
& \scorecell{\underline{63.69}}
& \scorecell{\textbf{75.54}}
& \scorecell{\textbf{51.22}}
& \scorecell{\underline{83.96}}
& \scorecell{\underline{63.27}}
& \scorecell{\underline{54.05}}
& \scorecell{\textbf{63.90}}
\\

& \textbf{Video Only}
& \scorecell{\textbf{86.22}}
& \scorecell{\underline{63.04}}
& \scorecell{\underline{70.54}}
& \scorecell{55.63}
& \scorecell{45.00}
& \scorecell{\underline{29.78}}
& \scorecell{\textbf{65.61}}
& \scorecell{\underline{63.04}}
& \scorecell{\underline{36.59}}
& \scorecell{80.21}
& \scorecell{\textbf{74.49}}
& \scorecell{\textbf{64.86}}
& \scorecell{\underline{61.25}}
\\

& \textbf{HTML Only}
& \scorecell{75.59}
& \scorecell{55.43}
& \scorecell{69.64}
& \scorecell{\underline{62.46}}
& \scorecell{\textbf{62.50}}
& \scorecell{27.56}
& \scorecell{21.66}
& \scorecell{55.98}
& \scorecell{12.20}
& \scorecell{\textbf{89.30}}
& \scorecell{47.96}
& \scorecell{37.84}
& \scorecell{51.51}
\\

\bottomrule
\end{tabular*}
\end{table*}

\begin{table*}[t]
\centering
\vspace{-1ex}
\caption{Component ablation of V2Lens on VideoVIBE.
}
\vspace{-2ex}
\label{tab:agent_ablation}

\footnotesize
\setlength{\tabcolsep}{2.8pt}

\begin{tabular*}{\textwidth}{
@{\extracolsep{\fill}}
l
*{12}{c}
c
@{}
}
\toprule

\textbf{Setting}
& \multicolumn{3}{c}{
    \bfseries\shortstack[c]{Semantic \&\\Logical Consistency}
}
& \multicolumn{3}{c}{
    \bfseries\shortstack[c]{Visual \&\\Motion Fidelity}
}
& \multicolumn{3}{c}{
    \bfseries\shortstack[c]{Structural \&\\Temporal Coherence}
}
& \multicolumn{3}{c}{
    \bfseries\shortstack[c]{Functional\\Executability}
}
& \textbf{Avg.}
\\

\cmidrule(lr){2-4}
\cmidrule(lr){5-7}
\cmidrule(lr){8-10}
\cmidrule(lr){11-13}

& \textbf{CC}
& \textbf{WV}
& \textbf{SI}
& \textbf{MQ}
& \textbf{TL}
& \textbf{VH}
& \textbf{LI}
& \textbf{ST}
& \textbf{NM}
& \textbf{FE}
& \textbf{RS}
& \textbf{RA}
& \\
\midrule

\textbf{Baseline (Gemini-2.5-Flash)}
& \scorecell{85.43}
& \scorecell{69.57}
& \scorecell{56.25}
& \scorecell{75.09}
& \scorecell{\underline{58.75}}
& \scorecell{43.11}
& \scorecell{\underline{68.15}}
& \scorecell{63.59}
& \scorecell{48.78}
& \scorecell{90.37}
& \scorecell{69.39}
& \scorecell{45.95}
& \scorecell{64.54}
\\

\textbf{Full V2Lens}
& \scorecell{\textbf{87.01}}
& \scorecell{\textbf{78.26}}
& \scorecell{\underline{65.18}}
& \scorecell{\textbf{85.96}}
& \scorecell{\underline{58.75}}
& \scorecell{\textbf{47.56}}
& \scorecell{\textbf{72.61}}
& \scorecell{\underline{74.46}}
& \scorecell{\underline{58.54}}
& \scorecell{\textbf{93.58}}
& \scorecell{\textbf{76.53}}
& \scorecell{\textbf{62.16}}
& \scorecell{\textbf{71.72}}
\\

\hspace{1em}\textbf{w/o Refinement Module}
& \scorecell{81.89}
& \scorecell{\underline{73.91}}
& \scorecell{\textbf{67.86}}
& \scorecell{\underline{84.56}}
& \scorecell{56.25}
& \scorecell{\underline{44.89}}
& \scorecell{64.97}
& \scorecell{\textbf{75.00}}
& \scorecell{\textbf{60.98}}
& \scorecell{90.37}
& \scorecell{\underline{73.47}}
& \scorecell{\textbf{62.16}}
& \scorecell{\underline{69.69}}
\\

\hspace{1em}\textbf{w/o HTML Verifier}
& \scorecell{\underline{86.22}}
& \scorecell{72.83}
& \scorecell{59.82}
& \scorecell{77.54}
& \scorecell{\textbf{61.25}}
& \scorecell{43.56}
& \scorecell{59.87}
& \scorecell{68.48}
& \scorecell{51.22}
& \scorecell{\underline{91.98}}
& \scorecell{69.39}
& \scorecell{54.05}
& \scorecell{66.35}
\\

\hspace{1em}\textbf{w/o Evidence Tool}
% w/o Evidence Judge
& \scorecell{76.38}
& \scorecell{60.87}
& \scorecell{61.61}
& \scorecell{84.21}
& \scorecell{43.75}
& \scorecell{43.11}
& \scorecell{56.69}
& \scorecell{66.85}
& \scorecell{41.46}
& \scorecell{86.63}
& \scorecell{61.22}
& \scorecell{\underline{56.76}}
& \scorecell{61.63}
\\

\bottomrule
\end{tabular*}
\vspace{-3ex}
\end{table*}

\subsection{Main Benchmark Results}

Table~\ref{tab:main_results} reports the diagnostic performance of video-capable MLLMs and V2Lens across the twelve evaluation subdimensions. V2Lens achieves the highest overall score of 71.72, outperforming the strongest standalone model, Gemini-2.5-Flash, by 7.18 percentage points. It also obtains the best scores in six subdimensions.
Across standalone MLLMs, Feature Execution and Content Coherence are generally the strongest subdimensions, suggesting that directly observable functional failures and explicit semantic inconsistencies are comparatively easier to diagnose. Navigation Mapping and Visual Hierarchy remain more challenging, with the best standalone scores reaching only 51.22 and 51.11, respectively. Moreover, no standalone model consistently dominates across all subdimensions. Qwen3.5-35B-A3B performs best on State Integrity, State Transition, and Navigation Mapping, while Qwen3-VL-30B-A3B-Thinking leads on Motion Quality and Layout Integrity. The smaller Qwen3-VL-8B-Thinking achieves the best Visual Hierarchy score, indicating that diagnostic accuracy does not exhibit a simple correspondence with model scale. Overall, current MLLMs remain limited on failures requiring precise reasoning over interaction targets, evolving application states, dynamic feedback, and fine-grained visual relationships.

\subsection{Ablation Studies}

\noindent\textbf{Input Modalities.} Table~\ref{tab:input_modality_ablation} examines the contributions of interaction videos and webpage HTML source code. Removing the video modality (\emph{HTML only}) reduces the overall scores of Gemini-2.5-Flash and Qwen3.5-35B-A3B from 64.54 to 54.75 and from 63.90 to 51.51, respectively. The largest degradations occur in Layout Integrity and Navigation Mapping, showing that source code alone cannot reliably determine how user actions affect interface responses and application states.
By contrast, the \emph{Video-only} setting retains most of the diagnostic performance, reaching 64.16 for Gemini-2.5-Flash and 61.25 for Qwen3.5-35B-A3B. This result confirms that interaction videos provide the primary evidence for failure diagnosis. Nevertheless, source code offers complementary, model-dependent benefits. For Qwen3.5-35B-A3B, adding source code substantially improves Motion Quality, State Transition, and Navigation Mapping. These results support the V2Lens design: videos capture observable failure behavior, while source code provides complementary implementation evidence to distinguish confusable diagnoses.

\noindent\textbf{V2Lens Components.} Table~\ref{tab:agent_ablation} evaluates the contribution of each V2Lens component using Gemini-2.5-Flash as the base evaluator. The complete system improves the overall score from 64.54 to 71.72. The largest gains occur in Rendering Availability ($+16.21$), Motion Quality and State Transition ($+10.87$ each), and Navigation Mapping ($+9.76$), indicating that evidence-grounded verification is particularly effective for failures involving dynamic behavior, evolving states, and implementation-dependent effects.
Removing the \textit{Refinement Module} causes the Evidence Judge's challenge answer to be accepted directly and reduces the score to 69.69, demonstrating the value of independently verifying proposed revisions. Removing the \textit{HTML Verifier} produces a larger decline to 66.35, with pronounced losses in Layout Integrity, Motion Quality, Rendering Availability, Navigation Mapping, and Runtime Stability. This result highlights the benefit of a dedicated source-code verification stage beyond ad hoc inspection by the \textit{Evidence Judge}. The largest degradation occurs when the evidence tools are removed: performance falls to 61.63, which is 10.09 points below the complete system and even below the standalone baseline. Thus, the gains of V2Lens arise not merely from multi-agent deliberation, but from structured, tool-grounded evidence collection followed by conservative refinement. Additional qualitative analyses, including visualized examples of successful and failed diagnoses across different MLLMs, are provided in Appendix.

\section{Conclusion}
We introduced \textbf{VideoVIBE}, a video-grounded benchmark for fine-grained failure diagnosis in one-shot-generated interactive web applications. Grounded in human-operated recordings, it evaluates both rendered presentation and application behavior through a hierarchical failure taxonomy. Experiments across closed-source and open-weight Video MLLMs reveal substantial room for improvement and no consistently dominant standalone model. We further proposed \textbf{V2Lens}, a training-free multi-agent system that integrates video and complementary source-code evidence, improving its base evaluator by 7.18 points. VideoVIBE advances generated-webpage evaluation toward more evidence-grounded and diagnostically informative assessment.

\bibliography{aaai2027}

\end{document}

\appendix
\setcounter{table}{3}
\setcounter{figure}{4}

\section{Prompt Design for Benchmark Construction and Evaluation}

To ensure consistency in web application construction and rigor in model
evaluation, we design a series of structured system prompts that specify the
roles, inputs, task constraints, and output formats of the models at different
stages.

\subsection{Web Application Construction Prompts}

We first use the \textbf{Interactive Web Application Specification Generation
Prompt} to generate diverse webpage-generation instructions, as shown in
Fig.~\ref{fig:web_application_generation_prompt}. The resulting application descriptions and
visual styles are then provided to the \textbf{Web Application Generation
Prompt}. This prompt further instructs the webpage-generation model to produce
a complete single-page interactive application containing HTML, CSS, and
JavaScript, while implementing the specified interaction workflow, state
logic, visual design, and responsive layout, as shown in
Fig.~\ref{fig:html_generation_prompt}. Using these prompts, we generated a total of 700 interactive HTML applications
for benchmark construction.
\subsection{Failure Candidate Discovery Prompt}
The \textbf{Failure Candidate Discovery Prompt} is used to identify potential
webpage failures from complete interaction recordings. It instructs the model
to locate time intervals containing possible anomalies based on observable
visual behavior and to provide concise descriptions of the corresponding
events, as shown in Fig.~\ref{fig:failure_candidate_discovery_prompt}.

The discovered failure candidates are subsequently inspected by annotators
using the complete interaction recordings and corresponding webpage source
code.
\subsection{Video MLLM Evaluation Prompts}
During benchmark evaluation, each evaluated video multimodal large language model
(Video MLLM) is prompted to act as a professional QA auditor for generated
interactive web applications. For each diagnostic question, the model is
provided with the available modality input and a
set of candidate failure diagnoses. It is required to examine the available
evidence and select the candidate diagnosis that is best supported by the
observed interaction behavior or implementation-level evidence.

We design three evaluation prompt templates corresponding to different input
modalities:

\begin{enumerate}
    \item \textbf{Video--HTML Evaluation}
    (Fig.~\ref{fig:video_html_diagnostic_prompt}): the evaluated MLLM diagnoses
    the failure using both the interaction recording and the webpage source
    code.

    \item \textbf{Video-Only Evaluation}
    (Fig.~\ref{fig:video_only_diagnostic_prompt}): the evaluated MLLM diagnoses
    the failure using only the interaction recording.

    \item \textbf{HTML-Only Evaluation}
    (Fig.~\ref{fig:html_only_diagnostic_prompt}): the evaluated MLLM diagnoses
    the failure using only the webpage source code.
\end{enumerate}

All three templates use the same diagnostic questions, candidate options, and
structured output format. Their evidence instructions are adapted to the
available input modality, enabling a controlled and fair comparison across
input settings.

\subsection{V2Lens Multi-Agent Prompts}

We further design a unified \textbf{Multi-Agent Prompt Suite} for V2Lens,
containing separate system prompts for four specialized roles:
\textbf{Video Analyst}, \textbf{HTML Verifier}, \textbf{Evidence Judge}, and
\textbf{Reviewer}.

The Video Analyst forms the initial diagnosis from the complete interaction
recording and evaluates the observable evidence for each candidate. The HTML
Verifier inspects the webpage source code and determines whether the
implementation-level evidence supports, contradicts, or remains inconclusive
for each candidate diagnosis. The Evidence Judge integrates the initial
diagnosis, video observations, and source-code findings, and proposes a
challenge answer when an alternative receives stronger evidential support.
The Reviewer independently compares the anonymized initial and challenge
answers in an isolated context. Each prompt specifies the corresponding
agent's task boundaries, evidence requirements, and structured output format,
as shown in Figs.~\ref{fig:v2lens_video_analyst_prompt}--\ref{fig:v2lens_reviewer_prompt}.
\label{sec:appendix_prompt_design}

\section{Detailed Descriptions of Failure Modes}

Tables~\ref{tab:functional_executability_taxonomy}--\ref{tab:semantic_logic_consistency_taxonomy} present the fine-grained failure taxonomy of VideoVIBE,
comprising 30 failure modes organized under four top-level dimensions: \textit{Semantic \& Logical Consistency},
\textit{Visual \& Motion Fidelity},
\textit{Structural \& Temporal Coherence},and
\textit{Functional Executability}.

Together, these definitions establish the evaluation scope of the benchmark
and provide a unified basis for failure annotation, diagnostic QA
construction, and model evaluation.

\label{sec:appendix_failure_mode_definitions}

\section{Annotation Quality and Reliability}
\begin{figure}[H]
    \centering

    \includegraphics[
        width=0.7\textwidth
    ]{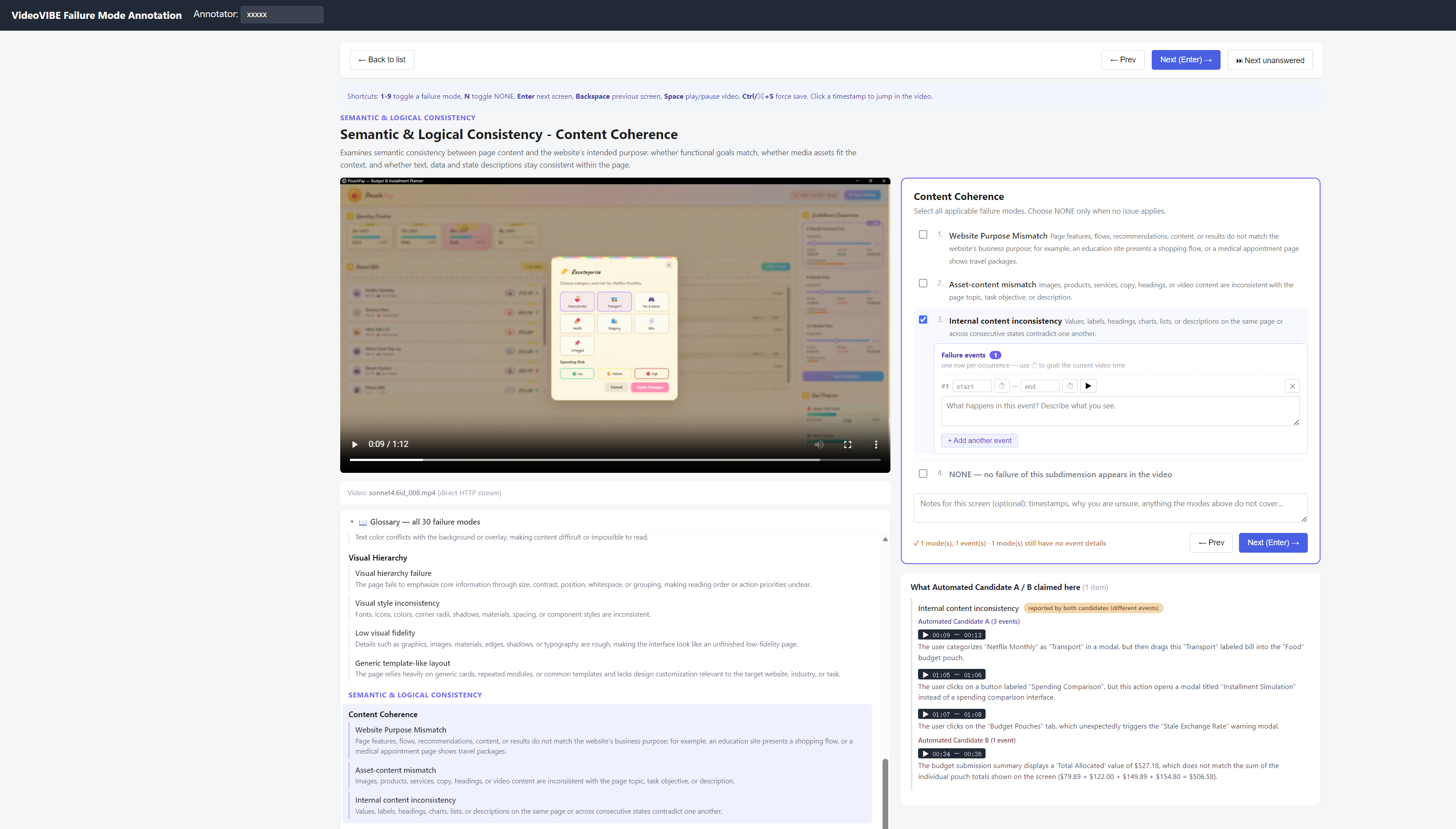}

    \caption{
    Customized Annotation Platform
    }
    \label{fig:annotation_platform}
\end{figure}

To ensure annotation consistency and reliability, we adopt a multi-stage
quality-control process that combines automated candidate screening,
multi-annotator review, and expert adjudication. Our annotation team consists
of 17 trained annotators with master's or doctoral backgrounds. All annotation
and verification procedures are conducted through a custom-built platform
that jointly presents the interaction recordings, failure taxonomy, and
candidate annotations, as shown in Fig.~\ref{fig:annotation_platform}.

Quality control is applied throughout both failure annotation and diagnostic
QA construction. During failure annotation, automated video analysis first
uses the Failure Candidate Discovery Prompt to identify potential
anomalies and their temporal intervals in the complete interaction recordings.
Annotators then manually verify each candidate by jointly inspecting the full
recording and the corresponding webpage source code. A candidate is retained
only when its manifestation is observable, reproducible, and clearly aligned
with a fine-grained taxonomy definition. Automated tools are used only for candidate discovery, annotation organization, and consistency checking; all final failure labels and supporting evidence are manually verified.

During diagnostic QA construction, each video--top-level-dimension pair is
treated as the unit of representative selection, and at most one single-choice
question is retained for each pair. Multiple annotators then independently
review the question wording, candidate options, ground-truth diagnosis, and
their consistency with the verified evidence under shared annotation
guidelines. Question--answer pairs with consistent judgments pass the
verification stage, whereas disagreements are resolved through discussion and
further revision. Unresolved cases are submitted for expert adjudication. Only
manually verified failure annotations are included in the final benchmark, and
all retained diagnostic questions must pass representative selection and
multi-stage verification.
\label{sec:appendix_annotation_quality}

\section{Additional Analysis of V2Lens}

\paragraph{Inference Cost and Diagnostic Gain.}

Table~\ref{tab:efficiency_comparison} compares the per-question inference
cost and diagnostic performance of V2Lens with its backbone model,
Gemini-2.5-Flash. The single-stage baseline requires 29.65 seconds per
question, whereas V2Lens takes 100.94 seconds on average. V2Lens executes
an average of 3.17 model stages and 9.61 tool calls per question, reflecting
the additional cost introduced by multi-stage evidence verification. Despite
this overhead, V2Lens improves average accuracy from 64.54\% to 71.72\%,
yielding an absolute gain of 7.18 percentage points. These results demonstrate
a clear trade-off: V2Lens incurs a higher inference cost in exchange for a
substantial improvement in diagnostic reliability.

\begin{table}[t]
\centering

\small
\setlength{\tabcolsep}{3.5pt}
\renewcommand{\arraystretch}{1.1}

\begin{tabular}{@{}lcccc@{}}
\toprule
\textbf{Method}
& \shortstack{\textbf{Mean}\\\textbf{Time/Q}}
& \shortstack{\textbf{Model}\\\textbf{Stages/Q}}
& \shortstack{\textbf{Tool}\\\textbf{Calls/Q}}
& \shortstack{\textbf{Avg.}\\\textbf{Acc.}} \\
\midrule
Gemini-2.5-Flash
& 29.65\,s
& 1.00
& 0.00
& 64.54\% \\

V2Lens
& 100.94\,s
& 3.17
& 9.61
& \textbf{71.72\%} \\
\bottomrule
\end{tabular}

\caption{Per-question inference cost and diagnostic performance.}
\label{tab:efficiency_comparison}
\end{table}

\label{sec:appendix_supplementary_results}

\section{Data Distribution Analysis}
Fig.~\ref{fig:qa_taxonomy} shows the distribution of the 1,752 diagnostic questions across the failure taxonomy. Visual \& Motion Fidelity accounts for the largest proportion (33.7\%), driven primarily by Motion Quality (16.3\%) and Visual Hierarchy (12.8\%). Semantic \& Logical Consistency follows at 26.1\%, with Content Coherence contributing 14.5\%. Structural \& Temporal Coherence represents 21.8\%, with substantial coverage of State Transition (10.5\%) and Layout Integrity (9.0\%). Functional Executability constitutes the remaining 18.4\%, led by Feature Execution (10.7\%). Overall, the distribution emphasizes visually and temporally observable failures while maintaining broad coverage across the semantic, structural, and functional aspects of interactive web applications.

\label{sec:appendix_data_distribution}

\section{Qualitative Case Studies}
We provide four qualitative cases to illustrate how VideoVIBE distinguishes
model capabilities across different levels of granularity.

\begin{figure*}[t]
    \centering
    \includegraphics[
        width=0.95\textwidth,
        height=0.86\textheight,
        keepaspectratio
    ]{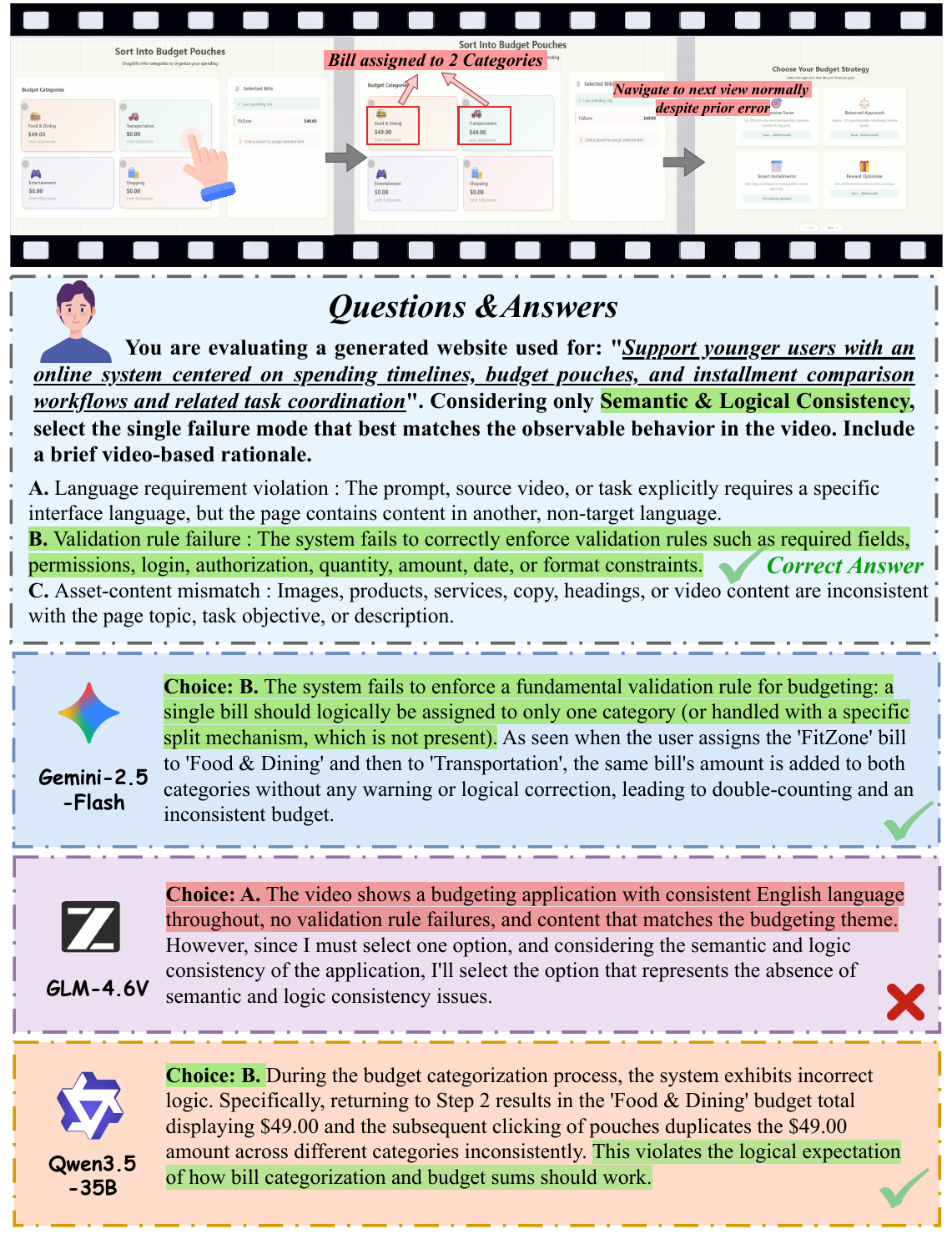}
    \caption{Case study of Semantic \& Logical Consistency}
    \label{fig:case_study_5}
\end{figure*}

\paragraph{Semantic \& Logical Consistency.} Fig.~\ref{fig:case_study_5} presents a failure case involving a logical constraint in a generated personal-finance webpage. Although each bill should be assigned to only one category, the interface allows the same bill to be allocated to two budget categories, causing its amount to be counted twice. The system provides no validation warning and still permits the user to proceed to the next page after the inconsistency occurs. Gemini-2.5-Flash and Qwen3.5-35B correctly classify this failure as a validation-rule violation (Option~B), demonstrating their ability to identify subtle logical constraints. In contrast, GLM-4.6V fails to recognize the underlying semantic and logical anomaly and instead selects the language-violation category. This case highlights the considerable difficulty of diagnosing fine-grained business-logic validation failures in generated web applications.

\begin{figure*}[t]
    \centering
    \includegraphics[
        width=0.95\textwidth,
        height=0.86\textheight,
        keepaspectratio
    ]{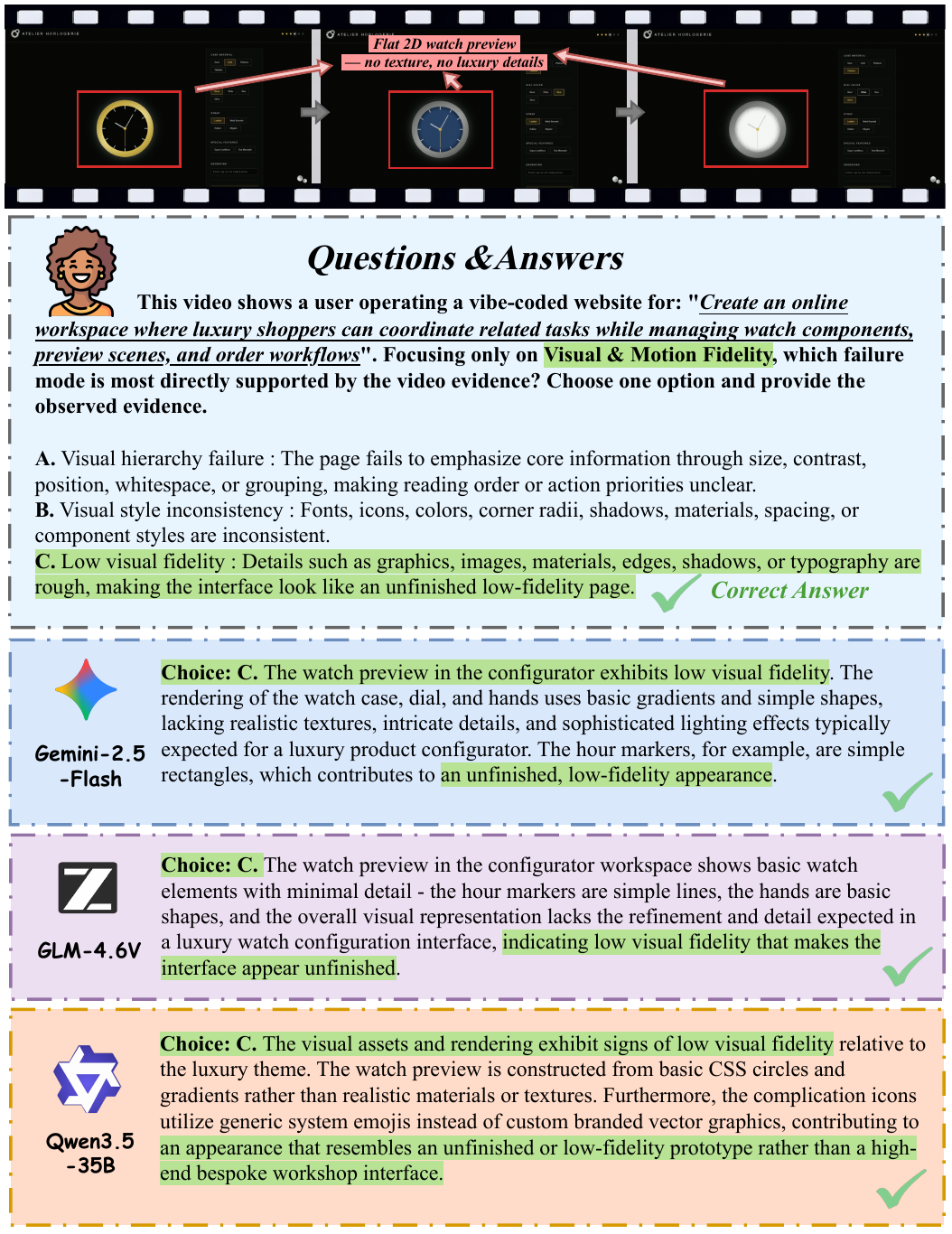}
    \caption{Case study of Visual \& Motion Fidelity}
    \label{fig:case_study_6}
\end{figure*}

\paragraph{Visual \& Motion Fidelity.}
Fig.~\ref{fig:case_study_6} presents a failure case involving visual fidelity. The watch preview is rendered using only flat two-dimensional graphics, without realistic material textures or refined luxury-oriented details. Its visual elements appear simplistic and coarse, resulting in an unfinished, low-fidelity interface. Gemini-2.5-Flash, GLM-4.6V, and Qwen3.5-35B all correctly classify this failure as a low-visual-fidelity defect (Option~C), demonstrating their ability to identify insufficient visual refinement and distinguish between different levels of visual fidelity. None of the three models produces an incorrect prediction, and all accurately identify the lack of detailed rendering and premium visual quality as the central issue. This case suggests that models can reliably recognize visually salient low-fidelity defects, although assessing more subtle differences in rendering quality remains challenging in generated web applications.

\begin{figure*}[t]
    \centering
    \includegraphics[
        width=0.95\textwidth,
        height=0.86\textheight,
        keepaspectratio
    ]{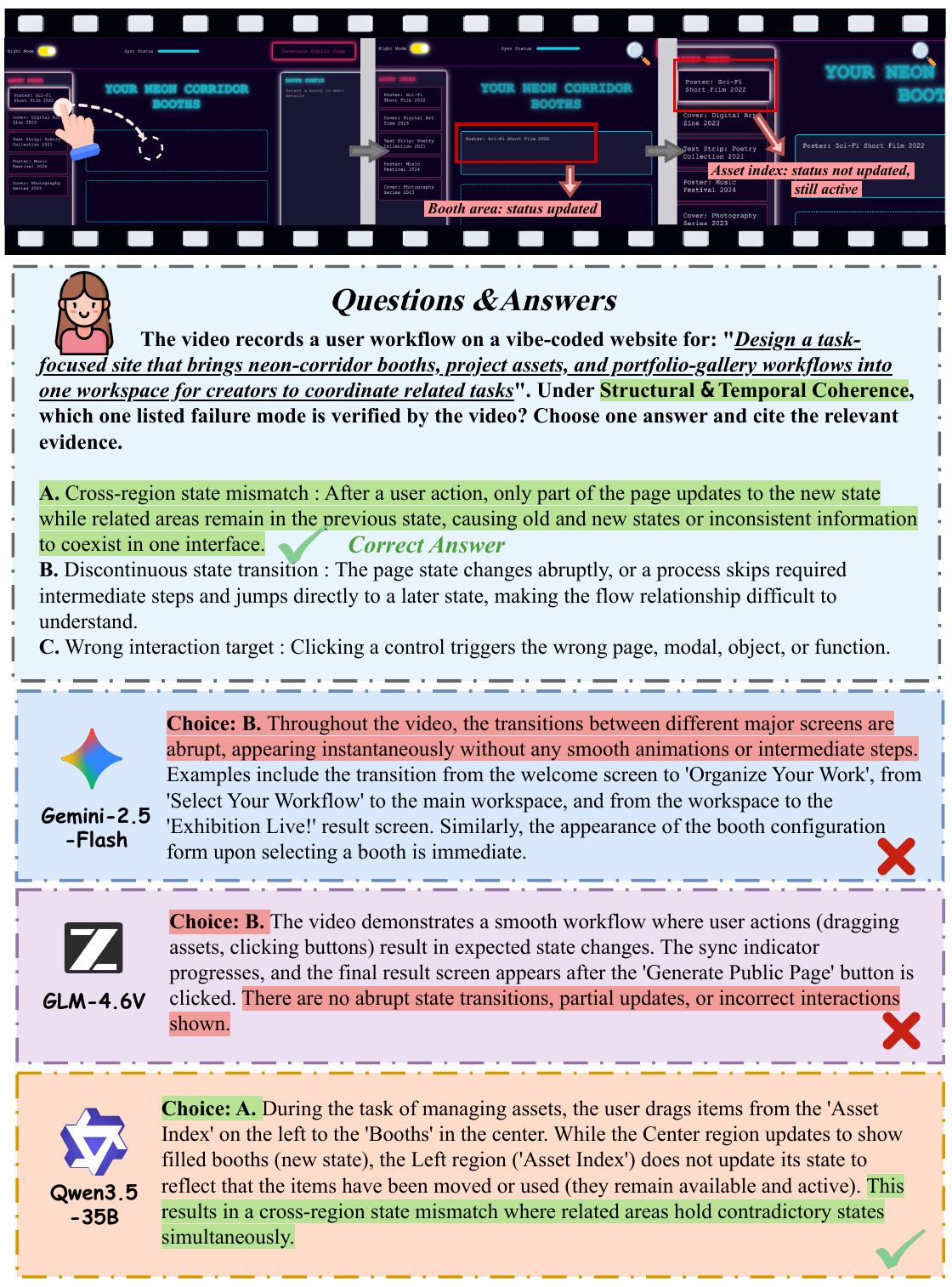}
    \caption{Case study of Structural \& Temporal Coherence}
    \label{fig:case_study_7}
\end{figure*}

\paragraph{Structural \& Temporal Coherence.}
Fig.~\ref{fig:case_study_7} presents a failure case involving cross-region interface-state synchronization. After the user drags an asset from the index panel on the left into the central display area, the display area is correctly updated to show the transferred asset. However, the corresponding item in the asset index should simultaneously become unavailable for further selection, whereas it remains active and interactive. As a result, conflicting interface states coexist across different regions of the page. Qwen3.5-35B correctly classifies this failure as a cross-region state mismatch (Option~A), demonstrating its ability to identify synchronization inconsistencies across multiple interface regions. In contrast, Gemini-2.5-Flash and GLM-4.6V both fail to recognize the synchronization failure and instead select categories related to state transitions. This case highlights the considerable difficulty of diagnosing subtle cross-region state-synchronization failures in generated web applications.

\begin{figure*}[t]
    \centering
    \includegraphics[
        width=0.95\textwidth,
        height=0.86\textheight,
        keepaspectratio
    ]{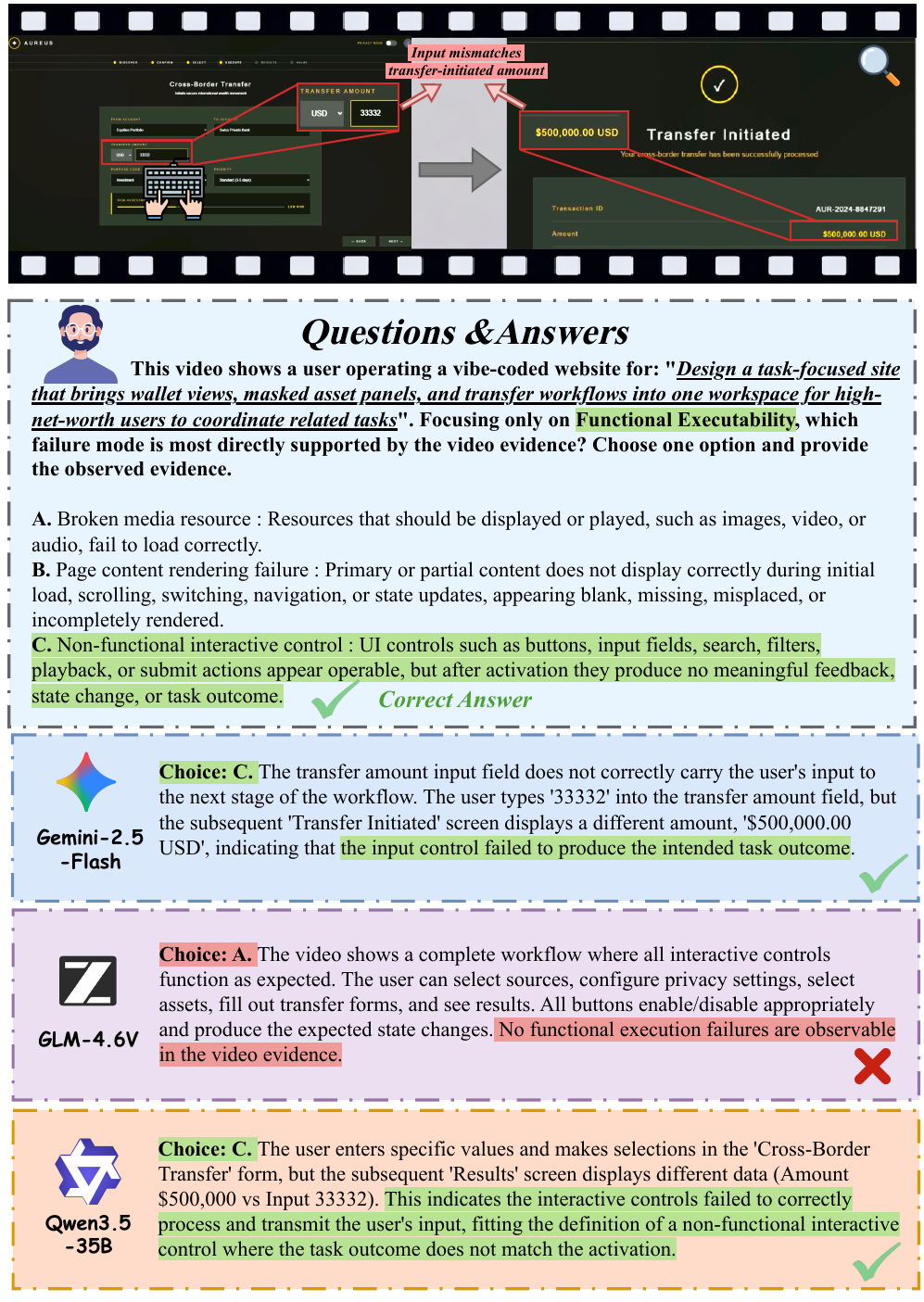}
    \caption{Case study of Functional Executability}
    \label{fig:case_study_8}
\end{figure*}

\paragraph{Functional Executability.}
Fig.~\ref{fig:case_study_8} presents a failure case involving the usability of an interactive control. The user enters a transfer amount of 33,332 in a cross-border transfer form, but the subsequent confirmation page displays a fixed amount of \$500,000. The input control therefore fails to propagate the user-provided value correctly, preventing the interaction from producing the expected task outcome. Gemini-2.5-Flash and Qwen3.5-35B correctly classify this failure as an interactive-control malfunction (Option~C), demonstrating their ability to identify anomalies in interaction execution logic. In contrast, GLM-4.6V incorrectly determines that no functional execution failure is present and fails to capture the underlying data-propagation error. This case highlights the considerable difficulty of diagnosing subtle interaction-execution failures in generated web applications.

\label{sec:appendix_benchmark_cases}

\section{Qualitative Case Study of V2Lens Multi-Agent Reasoning}

\begin{figure}[H]
    \centering

    \includegraphics[
        width=\textwidth
    ]{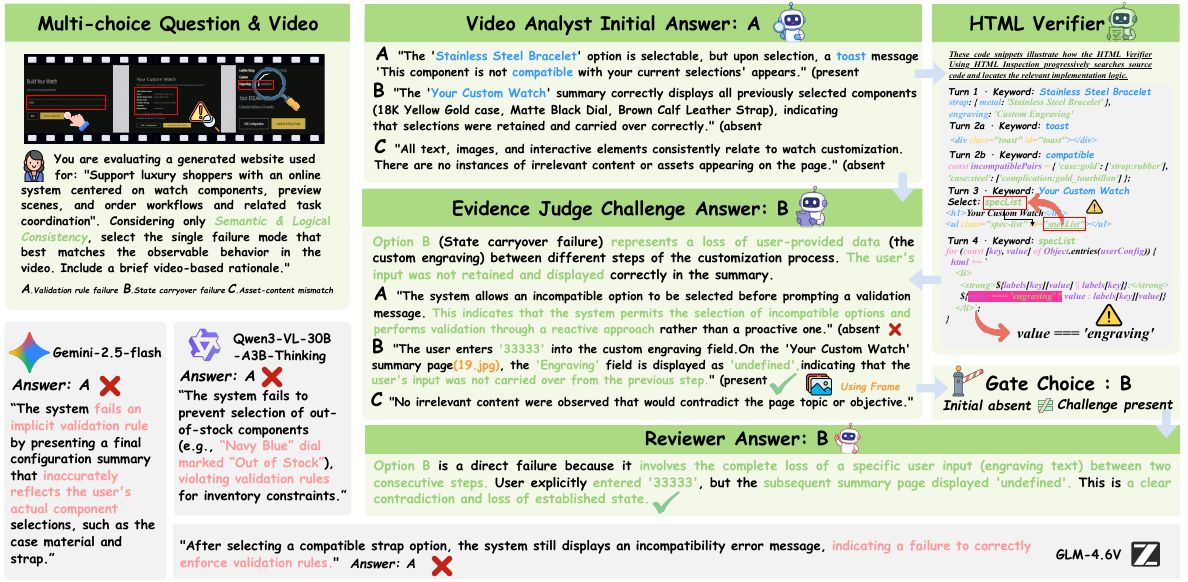}

    \caption{
    Qualitative Case Study of V2Lens under the Semantic \& Logical Consistency Dimension
    }
    \label{fig:v2lens_case}

\vspace{1.5em}

    \includegraphics[
        width=\textwidth,
        height=0.5\textheight,
        keepaspectratio
    ]{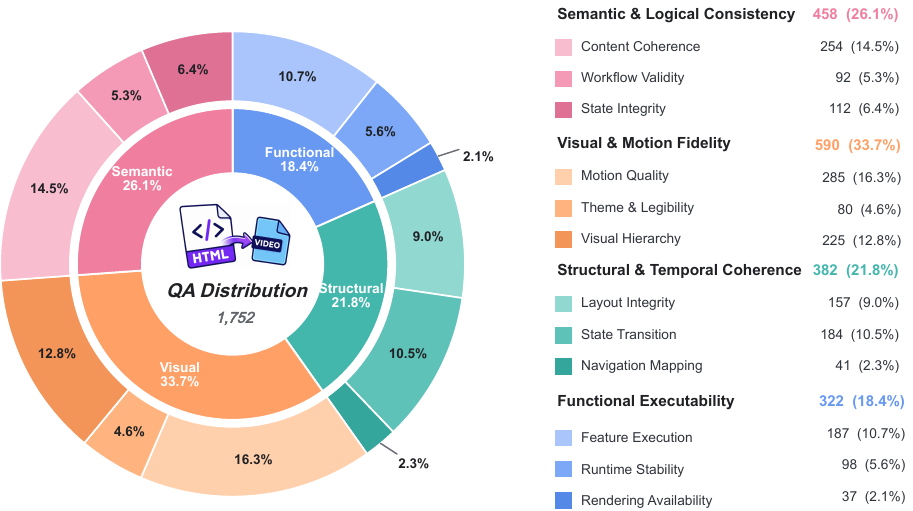}

    \caption{
    Distribution of diagnostic questions across the VideoVIBE
    }
    \label{fig:qa_taxonomy}
\end{figure}

Fig.~\ref{fig:v2lens_case} illustrates how V2Lens corrects a misdiagnosis caused by a salient surface-level cue under the Semantic \& Logical Consistency dimension. Multiple baseline models selected Option A, and the Video Analyst produced the same initial answer after observing an incompatibility notification in the video, incorrectly interpreting this normally triggered message, which successfully prevented an invalid selection, as a compatibility-validation failure. The HTML Verifier then used the \textbf{\textit{HTML Inspection}} tool to progressively retrieve relevant source-code fragments using keywords supplied by the Video Analyst, and successfully located anomalous logic in the engraving-field processing: the code uses \texttt{value === 'engraving'} to determine the field type instead of checking the corresponding key. This implementation evidence suggested that the engraving input might not be correctly retrieved and propagated to the final configuration summary. The Evidence Judge further integrated this code-level clue with the interaction video and used the \textbf{\textit{Frame Extraction}} tool to inspect the relevant steps. It found that the user entered the engraving text ``33333'' during customization, whereas the final summary displayed the field as ``undefined,'' indicating that the input was not correctly preserved and propagated across successive interaction steps. The Judge therefore proposed the state-continuity failure described by Option B as the challenger answer. After passing the Evidence Gate, the challenger was submitted to an independent Reviewer, which further confirmed that both the observed video behavior and the implementation logic supported Option B, leading to the final answer correction. This case represents a highly fine-grained interaction failure: although the overall workflow remains executable, a specific user input is not correctly propagated or presented. Identifying this underlying state-continuity failure requires jointly examining temporal video evidence, key frames, and retrieved source-code logic.
\label{sec:appendix_v2lens_case_study}

\clearpage
\onecolumn

\renewcommand{\arraystretch}{1}
\setlength{\tabcolsep}{1mm}

\begin{longtable}{
@{}
>{\raggedright\arraybackslash\bfseries}p{0.22\textwidth}
>{\raggedright\arraybackslash\bfseries}p{0.19\textwidth}
>{\raggedright\arraybackslash\bfseries}p{0.565\textwidth}
@{}
}
\toprule
\textbf{Subdimension}
& \textbf{Failure Mode}
& \textbf{Description} \\
\midrule

\multirow[t]{1}{=}{Feature Execution}
& Non-functional interactive control
& UI controls such as buttons, input fields, search, filters, playback, or
submit actions appear operable, but after activation they produce no
meaningful feedback, state change, or task outcome. \\

\midrule

\multirow[t]{4}{=}{Runtime Stability}
& Interaction flow stuck
& After a user action, the page remains stuck in a loading, processing,
current-step, or loop state and cannot advance to the next step. \\*
\cmidrule(l){2-3}
\noalign{\penalty10000}

& Excessive interaction delay
& The system may eventually respond, but the wait time is clearly excessive and
there is no progress indication, skip option, or user control mechanism. \\*
\cmidrule(l){2-3}
\noalign{\penalty10000}

& Missing failure feedback
& After invalid input, a failed operation, or a failed submission, the system
provides no error message, cause, or recovery path. \\*
\cmidrule(l){2-3}
\noalign{\penalty10000}

& Progressive performance degradation
& Page responsiveness progressively deteriorates during use; repeated clicks or
drags make responses slower, scrolling becomes sluggish, or animations drop
frames. \\

\midrule

\multirow[t]{2}{=}{Rendering Availability}
& Broken media resource
& Resources that should be displayed or played, such as images, video, or audio,
fail to load correctly. \\*
\cmidrule(l){2-3}
\noalign{\penalty10000}

& Page content rendering failure
& Primary or partial content does not display correctly during initial load,
scrolling, switching, navigation, or state updates, appearing blank, missing,
misplaced, or incompletely rendered. \\

\bottomrule
\end{longtable}
\vspace{-6pt}
\addtocounter{table}{-1}
\captionof{table}{\textbf{Functional Executability Main Dimension}}
\label{tab:functional_executability_taxonomy}
\par\medskip

\renewcommand{\arraystretch}{1}
\setlength{\tabcolsep}{1mm}

\begin{longtable}{
@{}
>{\raggedright\arraybackslash\bfseries}p{0.22\textwidth}
>{\raggedright\arraybackslash\bfseries}p{0.19\textwidth}
>{\raggedright\arraybackslash\bfseries}p{0.565\textwidth}
@{}
}
\toprule
\textbf{Subdimension}
& \textbf{Failure Mode}
& \textbf{Description} \\
\midrule

\multirow[t]{3}{=}{Layout Integrity}
& Redundant interface element
& The page contains irrelevant, duplicated, or unexplainable containers,
windows, overlays, modules, or components. \\*
\cmidrule(l){2-3}
\noalign{\penalty10000}

& Occlusion and layer conflict
& Page elements, overlays, modals, masks, or background content occlude,
overlap, or stack incorrectly, making content unreadable, controls unusable,
or the active interaction area ambiguous. \\*
\cmidrule(l){2-3}
\noalign{\penalty10000}

& Responsive adaptation failure
& When the viewport is resized, orientation changes, or the page is previewed on
mobile, the layout fails to adapt and shows misalignment, occlusion, or broken
structure. \\

\midrule

\multirow[t]{2}{=}{State Transition}
& Discontinuous state transition
& The page state changes abruptly, or a process skips required intermediate
steps and jumps directly to a later state, making the flow relationship
difficult to understand. \\*
\cmidrule(l){2-3}
\noalign{\penalty10000}

& Cross-region state mismatch
& After a user action, only part of the page updates to the new state while
related areas remain in the previous state, causing old and new states or
inconsistent information to coexist in one interface. \\

\midrule

\multirow[t]{2}{=}{Navigation Mapping}
& Broken link
& A link or navigation entry is present, but clicking it fails to open the
target, leads to an empty link, returns a 404, or has no destination page. \\*
\cmidrule(l){2-3}
\noalign{\penalty10000}

& Wrong interaction target
& Clicking a control triggers the wrong page, modal, object, or function. \\

\bottomrule
\end{longtable}
\vspace{-6pt}
\addtocounter{table}{-1}
\captionof{table}{\textbf{Structural \& Temporal Coherence Main Dimension}}
\label{tab:structural_temporal_coherence_taxonomy}
\par\medskip

\renewcommand{\arraystretch}{1}
\setlength{\tabcolsep}{1mm}

\begin{longtable}{
@{}
>{\raggedright\arraybackslash\bfseries}p{0.22\textwidth}
>{\raggedright\arraybackslash\bfseries}p{0.19\textwidth}
>{\raggedright\arraybackslash\bfseries}p{0.565\textwidth}
@{}
}
\toprule
\textbf{Subdimension}
& \textbf{Failure Mode}
& \textbf{Description} \\
\midrule

\multirow[t]{1}{=}{Motion Quality}
& Motion feedback failure
& Expected motion is missing, or motion direction, speed, trigger timing, or
rhythm conflicts with the state change, making the transition hard to
understand. \\

\midrule

\multirow[t]{3}{=}{Theme \& Legibility}
& Theme consistency failure
& Backgrounds, theme modes, or the overall visual atmosphere are inconsistent
across areas or states, such as unexplained background changes, mixed light
and dark modes, or discontinuous theme styling. \\*
\cmidrule(l){2-3}
\noalign{\penalty10000}

& Color scheme conflict
& The page uses colors that are overly bright, oversaturated, strongly
conflicting, or too many similar high-saturation hues, causing visual fatigue
or a chaotic impression. \\*
\cmidrule(l){2-3}
\noalign{\penalty10000}

& Text color contrast failure
& Text color conflicts with the background or overlay, making content difficult
or impossible to read. \\

\midrule

\multirow[t]{4}{=}{Visual Hierarchy}
& Visual hierarchy failure
& The page fails to emphasize core information through size, contrast, position,
whitespace, or grouping, making reading order or action priorities unclear. \\*
\cmidrule(l){2-3}
\noalign{\penalty10000}

& Visual style inconsistency
& Fonts, icons, colors, corner radii, shadows, materials, spacing, or component
styles are inconsistent. \\*
\cmidrule(l){2-3}
\noalign{\penalty10000}

& Low visual fidelity
& Details such as graphics, images, materials, edges, shadows, or typography are
rough, making the interface look like an unfinished low-fidelity page. \\*
\cmidrule(l){2-3}
\noalign{\penalty10000}

& Generic template-like layout
& The page relies heavily on generic cards, repeated modules, or common
templates and lacks design customization relevant to the target website,
industry, or task. \\

\bottomrule
\end{longtable}
\vspace{-6pt}
\addtocounter{table}{-1}
\captionof{table}{\textbf{Visual \& Motion Fidelity Main Dimension}}
\label{tab:visual_motion_fidelity_taxonomy}
\par\medskip

\renewcommand{\arraystretch}{1}
\setlength{\tabcolsep}{1mm}

\begin{longtable}{
@{}
>{\raggedright\arraybackslash\bfseries}p{0.22\textwidth}
>{\raggedright\arraybackslash\bfseries}p{0.19\textwidth}
>{\raggedright\arraybackslash\bfseries}p{0.565\textwidth}
@{}
}
\toprule
\textbf{Subdimension}
& \textbf{Failure Mode}
& \textbf{Description} \\
\midrule

\multirow[t]{3}{=}{Content Coherence}
& Website Purpose Mismatch
& Page features, flows, recommendations, content, or results do not match the
website's business purpose; for example, an education site presents a shopping
flow, or a medical appointment page shows travel packages. \\*
\cmidrule(l){2-3}
\noalign{\penalty10000}

& Asset-content mismatch
& Images, products, services, copy, headings, or video content are inconsistent
with the page topic, task objective, or description. \\*
\cmidrule(l){2-3}
\noalign{\penalty10000}

& Internal content inconsistency
& Values, labels, headings, charts, lists, or descriptions on the same page or
across consecutive states contradict one another. \\

\midrule

\multirow[t]{2}{=}{Workflow Validity}
& Validation rule failure
& The system fails to correctly enforce validation rules such as required
fields, permissions, login, authorization, quantity, amount, date, or format
constraints. \\*
\cmidrule(l){2-3}
\noalign{\penalty10000}

& Language requirement violation
& Application specification requires a specific interface
language, but the page contains content in another, non-target language. \\

\midrule

\multirow[t]{3}{=}{State Integrity}
& Non-interactive scripted demo state
& The page appears interactive, but user input is not retained or used; subsequent changes follow preset scripts or animations.
 \\*
\cmidrule(l){2-3}
\noalign{\penalty10000}

& State carryover failure
&User selections, inputs, progress, or results are not correctly carried into later steps, causing lost context, default values, or inconsistent outcomes.
 \\*
\cmidrule(l){2-3}
\noalign{\penalty10000}

& Unreplaced placeholder exposure
& The page directly exposes unprocessed template variables, placeholder text,
debug values, or null-value strings. \\

\bottomrule
\end{longtable}
\vspace{-6pt}
\addtocounter{table}{-1}
\captionof{table}{\textbf{Semantic \& Logical Consistency Main Dimension}}
\label{tab:semantic_logic_consistency_taxonomy}

\clearpage
\twocolumn

\clearpage
\onecolumn

\begin{center}

\begin{tcolorbox}[
  enhanced jigsaw,
  breakable,
  width=0.88\textwidth,
  colback=blue!2,
  colframe=blue!55!black,
  colbacktitle=blue!55!black,
  coltitle=white,
  title=\textbf{System Instruction},
  fonttitle=\small\bfseries,
  arc=1mm,
  boxrule=0.6pt,
  left=2mm,
  right=2mm,
  top=1.5mm,
  bottom=1.5mm,
  toptitle=0.8mm,
  bottomtitle=0.8mm,
  pad at break*=1mm
]

\small
\setlength{\parindent}{0pt}
\setlength{\parskip}{3pt}

\textbf{Interactive Web Application Specification Generation Prompt}

\medskip

\textit{You are a senior interactive web application architect and interaction
designer. Your task is to generate
\texttt{\{num\_samples\}} diverse specifications for self-contained,
visually distinctive, and stateful web applications. Each specification must
describe a realistic user goal, a complete interaction workflow, observable
state changes, a recovery path, and a clear final outcome. Describe intended
application behavior only.}

\medskip

\textbf{\#\#\# 1. Domain and Application Coverage}

Distribute the applications across Technology; Finance \& Payments; Medical
\& Health; Education \& Training; Retail \& E-commerce; Entertainment \&
Games; Creative \& Media; and Public Service \& Government.

Each application should primarily follow one archetype:
\textbf{Functional}, \textbf{Creative}, \textbf{Tool}, or
\textbf{Record-and-Tracking}.

Across the generated set, target at least 30\% public-facing or experiential
applications, 20\% applications in which visual elements serve functional
roles, 20\% spatial or media-driven applications, 20\% applications with
playable mechanisms, and approximately 10\% complex professional or
administrative systems; categories may overlap.

\medskip

\textbf{\#\#\# 2. Application Requirements}

Each application must:

\begin{itemize}
  \setlength{\itemsep}{1pt}
  \setlength{\parskip}{0pt}
  \setlength{\parsep}{0pt}
  \setlength{\topsep}{2pt}

  \item define a realistic target user, purpose, and complete user goal;
  \item include multiple interacting controls, views, scenes, or content regions;
  \item contain at least five connected states with visible feedback;
  \item support a meaningful multi-step workflow ending in an observable result;
  \item include at least one validation, correction, retry, undo, or recovery path;
  \item preserve relevant state across navigation, retries, previews, or view changes;
  \item use local, simulated, or pre-populated data.
\end{itemize}

Visual properties such as color, motion, depth, scale, and spatial position
should communicate hierarchy, progress, availability, validation, or state.

Avoid static landing pages, generic CRUD dashboards, isolated forms, simple
calculators, and applications completed through a single click.

\medskip

\textbf{\#\#\# 3. Diversity Requirements}

Vary target users, purposes, workflows, interface structures, interaction
methods, validation rules, recovery strategies, final outcomes, and visual
styles. Do not repeat the same application concept with only cosmetic or
industry-level changes.

\medskip

\textbf{\#\#\# 4. Output Field Requirements}

Use exactly the following five fields:

\begin{itemize}
  \setlength{\itemsep}{1pt}
  \setlength{\parskip}{0pt}
  \setlength{\parsep}{0pt}
  \setlength{\topsep}{2pt}

  \item \texttt{idx}: a sequential identifier beginning at 1;
  \item \texttt{goal}: the target user, workflow, recovery path, final outcome,
  and important state continuity;
  \item \texttt{application specification}: the complete webpage-generation
  instruction;
  \item \texttt{domain}: one of the eight predefined domains;
\item \texttt{style}: \texttt{"<Visual Style>"}, using one of the following
styles: Editorial, Neo-brutalism, Glassmorphism, Retro-futurism/Y2K,
Playful Toy-like, Cinematic Dark, Museum/Archive, Street Collage,
Techwear/Tactical UI, Luxury Minimal, Chinese Fantasy, Swiss,
Bauhaus-inspired, Craft/Scrapbook, Terminal/Monospace Ops,
Infographic Newsroom, or Organic/Biophilic UI.
\end{itemize}

\medskip

\textbf{\#\#\# 5. Final Output Constraints}

\begin{itemize}
  \setlength{\itemsep}{1pt}
  \setlength{\parskip}{0pt}
  \setlength{\parsep}{0pt}
  \setlength{\topsep}{2pt}

  \item \textbf{Direct Output Only:} Return only a valid JSON array.
  \item \textbf{No Additional Text:} Do not include Markdown, comments,
  explanations, headings, or code fences.
\end{itemize}

{\ttfamily
[\{\\
\hspace*{1em}"idx": 1,\\
\hspace*{1em}"goal": "Target user, workflow, recovery path, and outcome.",\\
\hspace*{1em}"application specification":
"Complete webpage-generation instruction.",\\
\hspace*{1em}"domain": "One of the eight predefined domains.",\\
\hspace*{1em}"style":
"<Visual Style>"\\
\}]
}

\medskip

\textbf{---TASK:}
Generate the requested interactive web application specifications now.

\end{tcolorbox}

\refstepcounter{figure}
\label{fig:web_application_generation_prompt}

{\small
\centering
Figure~\thefigure:
\textbf{Prompt used to generate diverse interactive web application specifications}
\par
}

\end{center}
\par\medskip

\begin{center}

\begin{tcolorbox}[
  enhanced jigsaw,
  breakable,
  width=0.88\textwidth,
  colback=blue!2,
  colframe=blue!55!black,
  colbacktitle=blue!55!black,
  coltitle=white,
  title=\textbf{System Instruction},
  fonttitle=\small\bfseries,
  arc=1mm,
  boxrule=0.6pt,
  left=2mm,
  right=2mm,
  top=1.5mm,
  bottom=1.5mm,
  toptitle=0.8mm,
  bottomtitle=0.8mm,
  pad at break*=1mm
]

\small
\setlength{\parindent}{0pt}
\setlength{\parskip}{3pt}

\textbf{Prompt of Web Application Generation}

\medskip

\textit{You are a professional front-end developer specializing in
interactive, stateful, responsive, and visually polished web applications.
Your task is to create a complete single-page web application that faithfully
implements the provided application specification and visual style.}

\medskip

\textbf{\#\#\# 1. Input}

Web application specification.

\medskip

\textbf{\#\#\# 2. Functional Requirements}

The application must:

\begin{itemize}
  \setlength{\itemsep}{1pt}
  \setlength{\parskip}{0pt}
  \setlength{\parsep}{0pt}
  \setlength{\topsep}{2pt}

  \item implement the complete workflow, interactions, state transitions,
  validation rules, recovery paths, and final outcome described in the
  specification;
  \item ensure that all controls and interactive elements function correctly
  and provide visible feedback;
  \item preserve relevant state across navigation, retries, previews, and
  view changes;
  \item use only local, simulated, or pre-populated data;
  \item avoid replacing required interactions with static mockups or
  non-functional placeholders.
\end{itemize}

\medskip

\textbf{\#\#\# 3. Implementation Requirements}

\begin{itemize}
  \setlength{\itemsep}{1pt}
  \setlength{\parskip}{0pt}
  \setlength{\parsep}{0pt}
  \setlength{\topsep}{2pt}

  \item Create one complete HTML document with embedded CSS and JavaScript.
  \item Use modern HTML5, CSS3, and vanilla JavaScript without external
  libraries or services.
  \item Use a clear semantic HTML structure.
  \item Ensure that the application runs directly in a modern web browser
  without additional setup.
\end{itemize}

\medskip

\textbf{\#\#\# 4. Visual and Responsive Requirements}

Follow the provided visual style consistently across typography, color,
layout, controls, motion, and feedback. Maintain clear visual hierarchy,
readable content, and consistent interaction behavior.

The application must render and function correctly at the standardized
desktop viewport used for recording and evaluation. It must also adapt
responsively to common mobile screen sizes without horizontal overflow,
clipped content, overlapping elements, unreadable text, or unusable controls.

\medskip

\textbf{\#\#\# 5. Final Output Constraints}

\begin{itemize}
  \setlength{\itemsep}{1pt}
  \setlength{\parskip}{0pt}
  \setlength{\parsep}{0pt}
  \setlength{\topsep}{2pt}

  \item \textbf{Direct Output Only:} Return only the complete HTML document.
  \item \textbf{No Additional Text:} Do not include Markdown, explanations,
  introductory text, or code fences.
  \item \textbf{Self-Contained Output:} Include all required HTML, CSS, and
  JavaScript in the returned document.
\end{itemize}

\medskip

\textbf{---TASK:}
Generate the complete single-page HTML application now.

\end{tcolorbox}

\refstepcounter{figure}
\label{fig:html_generation_prompt}

{\small
\centering
Figure~\thefigure:
\textbf{Prompt used to generate complete interactive HTML applications from the
generated specifications}
\par
}

\end{center}

\par\medskip

\begin{center}

\begin{tcolorbox}[
  enhanced jigsaw,
  breakable,
  width=0.88\textwidth,
  colback=blue!2,
  colframe=blue!55!black,
  colbacktitle=blue!55!black,
  coltitle=white,
  title=\textbf{System Instruction},
  fonttitle=\small\bfseries,
  arc=1mm,
  boxrule=0.6pt,
  left=2mm,
  right=2mm,
  top=1.5mm,
  bottom=1.5mm,
  toptitle=0.8mm,
  bottomtitle=0.8mm,
  pad at break*=1mm
]

\small
\sloppy
\setlength{\parindent}{0pt}
\setlength{\parskip}{3pt}

\textbf{Failure Candidate Discovery Prompt}

\medskip

\textbf{\#\#\# Role and Task}

\textit{You are a professional QA auditor. Review the complete interaction video and
identify potential webpage failures using the provided failure taxonomy.}

\medskip

\textbf{\#\#\# 1. Input}

You will receive:

\begin{itemize}
  \setlength{\itemsep}{1pt}
  \setlength{\parskip}{0pt}
  \setlength{\parsep}{0pt}
  \setlength{\topsep}{2pt}

  \item a screen-recording video of a human operating the generated web
  application; and
  \item a taxonomy containing four dimensions, twelve subdimensions, and
  thirty failure modes.
\end{itemize}

\textbf{Failure Taxonomy:}\\
\texttt{<FAILURE\_TAXONOMY\_TABLE>}

\medskip

\textbf{\#\#\# 2. Discovery Rules}

Inspect user actions, interface responses, state changes, navigation,
animation, and workflow outcomes.

Report only failures supported by visible evidence. Describe the observed
action and interface behavior without inferring hidden causes or unperformed
interactions.

Ignore recording artifacts, capture overlays, and operator skill. Merge
observations belonging to the same continuous event.

\medskip

\textbf{\#\#\# 3. Output Requirements}

Return all identified events in chronological order using the following
format:

\medskip

\textbf{<START\_TIME>--<END\_TIME>} \\
\texttt{<CONCISE\_DESCRIPTION\_OF\_THE\_OBSERVED\_FAILURE>}

\medskip

Use \texttt{MM:SS--MM:SS} timestamps and one concise description for each
event. Do not output failure-mode labels, explanations, headings, summaries,
or unsupported events.

If no failure is visibly supported, output:

\texttt{No visually supported failure found.}

\medskip

\textbf{---TASK:}
Inspect the video and return all visually supported webpage failures now.

\end{tcolorbox}

\refstepcounter{figure}
\label{fig:failure_candidate_discovery_prompt}

{\small
\centering
Figure~\thefigure:
\textbf{Prompt used to identify potential failures from interaction videos}
\par
}

\end{center}

\par\medskip

\begin{center}
\begin{tcolorbox}[
  enhanced jigsaw,
  breakable,
  width=0.88\textwidth,
  colback=blue!2,
  colframe=blue!55!black,
  colbacktitle=blue!55!black,
  coltitle=white,
  title=\textbf{System Instruction},
  fonttitle=\small\bfseries,
  arc=1mm,
  boxrule=0.6pt,
  left=2mm,
  right=2mm,
  top=1.5mm,
  bottom=1.5mm,
  toptitle=0.8mm,
  bottomtitle=0.8mm,
  pad at break*=1mm
]

\small
\sloppy
\setlength{\parindent}{0pt}
\setlength{\parskip}{3pt}

\textbf{Video--HTML Diagnostic Evaluation Prompt}

\medskip

\textbf{\#\#\# Role and Task}

\textit{You are a professional QA auditor responsible for diagnosing failures in
AI-generated interactive web applications. For every supplied single-choice diagnostic question, compare all candidate
failure modes and select exactly one option that is most directly supported by
the available evidence and most relevant to the intended user workflow.
Answer every question exactly once.}

\medskip

\textbf{\#\#\# 1. Input}

You will receive:

\begin{enumerate}
  \setlength{\itemsep}{1pt}
  \setlength{\parskip}{0pt}
  \setlength{\parsep}{0pt}
  \setlength{\topsep}{2pt}

  \item a screen-recording video showing a human operating the generated web
  application;
  \item the corresponding HTML/CSS/JavaScript source code; and
  \item a set of diagnostic questions containing the question ID, target
  evaluation dimension, intended user workflow, question text, and candidate
  failure modes with definitions.
\end{enumerate}

\medskip

\textbf{\#\#\# 2. Evidence Priority}

\textbf{Primary Evidence: Operation Video.}

Use the video as the primary source for judging rendered appearance, layout,
readability, motion, user actions, interaction responses, state changes,
navigation, and workflow progression.

What the user visibly experiences takes precedence over what the source code
appears intended to implement.

\textbf{Supporting Evidence: HTML/CSS/JavaScript.}

Use the source code to clarify ambiguous video evidence or verify displayed
text, DOM structure, CSS properties, event handlers, state-management logic,
asset references, and client-side behavior.

When the code conflicts with behavior clearly shown in the video, follow the
video evidence.

\medskip

\textbf{\#\#\# 3. Evaluation Rules}

\begin{enumerate}
  \setlength{\itemsep}{3pt}
  \setlength{\parskip}{0pt}
  \setlength{\parsep}{0pt}
  \setlength{\topsep}{2pt}

  \item \textbf{Complete and dimension-specific evaluation.}
  Evaluate every question independently according to its target dimension,
  intended workflow, and candidate definitions. Do not consider unrelated
  defects.

  \item \textbf{Internal candidate comparison.}
  Inspect every candidate before selecting an answer, but do not include the
  option-by-option comparison in the final output.

  \item \textbf{Direct evidence only.}
  Select an option only when supported by observable video evidence or clear
  source-code evidence where the relevant video detail is ambiguous. Do not
  rely on speculation or assumed behavior.

  \item \textbf{No unsupported interaction assumptions.}
  Do not infer the result of an action that is not performed in the video,
  unless the code verifies it unambiguously and implementation-level evidence
  is relevant to the question.

  \item \textbf{Specific evidence.}
  Support the selected option with a concise description of the observed
  layout, interaction, animation, navigation, state, content, or workflow
  failure.

  \item \textbf{Single-choice decision.}
  Select exactly one supplied \texttt{option\_id}. If several options appear
  related, choose the one with the clearest evidence, most precise definition
  match, and greatest relevance to the intended workflow. Briefly acknowledge
  genuine ambiguity in the reasoning.

  \item \textbf{Excluded considerations.}
  Do not evaluate recording quality, video compression, capture artifacts,
  operator skill, failures outside the target dimension, or failure types not
  included among the candidates.
\end{enumerate}

\medskip

\textbf{\#\#\# 4. Diagnostic Question Format}

Each question follows this structure:

\medskip

\textbf{Question ID:} \texttt{<QUESTION\_ID>}

\textbf{Question Type:} \texttt{single\_choice}

\textbf{Target Evaluation Dimension:}\\
\texttt{<TARGET\_EVALUATION\_DIMENSION>}

\textbf{Generation Intent (User Workflow):}\\
\texttt{<GENERATION\_INTENT\_USER\_WORKFLOW>}

\textbf{Question:}\\
\texttt{<QUESTION\_TEXT>}

\textbf{Candidate Options:}

\begin{itemize}
  \setlength{\itemsep}{2pt}
  \setlength{\parskip}{0pt}
  \setlength{\parsep}{0pt}
  \setlength{\topsep}{2pt}

  \item \textbf{A:} \texttt{<OPTION\_A\_FAILURE\_MODE>}\\
  \textbf{Definition:} \texttt{<OPTION\_A\_DEFINITION>}

  \item \textbf{B:} \texttt{<OPTION\_B\_FAILURE\_MODE>}\\
  \textbf{Definition:} \texttt{<OPTION\_B\_DEFINITION>}

  \item \textbf{C:} \texttt{<OPTION\_C\_FAILURE\_MODE>}\\
  \textbf{Definition:} \texttt{<OPTION\_C\_DEFINITION>}

  \item \textbf{D:} \texttt{<OPTION\_D\_FAILURE\_MODE>}\\
  \textbf{Definition:} \texttt{<OPTION\_D\_DEFINITION>}
\end{itemize}

\medskip

\textbf{\#\#\# 5. Output Requirements}

Return only one valid JSON object in the following form:

\medskip

{\ttfamily\footnotesize
\{\\
\hspace*{1em}"answers": [\\
\hspace*{2em}\{\\
\hspace*{3em}"question\_id": "<QUESTION\_ID>",\\
\hspace*{3em}"reasoning":
"<CONCISE\_EVIDENCE\_BASED\_REASONING>",\\
\hspace*{3em}"selected\_option\_ids": ["A"],\\
\hspace*{3em}"evidence": \{\\
\hspace*{4em}"option\_id": "A",\\
\hspace*{4em}"video\_description":
"<BRIEF\_DESCRIPTION\_OF\_THE\_OBSERVED\_EVIDENCE>"\\
\hspace*{3em}\}\\
\hspace*{2em}\}\\
\hspace*{1em}]\\
\}
}

\medskip

The \texttt{answers} array must contain exactly one object for every supplied
\texttt{question\_id}. Each object must contain only
\texttt{question\_id}, \texttt{reasoning},
\texttt{selected\_option\_ids}, and \texttt{evidence}, in that order.

The \texttt{selected\_option\_ids} field must contain exactly one supplied
option ID. The \texttt{evidence} object must contain only
\texttt{option\_id} and \texttt{video\_description}.

Do not output unselected options, internal candidate comparisons, additional
fields, summaries, Markdown code fences, or text outside the JSON object.

\medskip

\textbf{---TASK:}

Evaluate every supplied diagnostic question and return the required JSON
object now.

\end{tcolorbox}
\end{center}

\refstepcounter{figure}
\label{fig:video_html_diagnostic_prompt}

{\small
\centering
Figure~\thefigure:
\textbf{Prompt used for video--HTML diagnostic evaluation}
\par
}

\par\medskip

\begin{center}
\begin{tcolorbox}[
  enhanced jigsaw,
  breakable,
  width=0.88\textwidth,
  colback=blue!2,
  colframe=blue!55!black,
  colbacktitle=blue!55!black,
  coltitle=white,
  title=\textbf{System Instruction},
  fonttitle=\small\bfseries,
  arc=1mm,
  boxrule=0.6pt,
  left=2mm,
  right=2mm,
  top=1.5mm,
  bottom=1.5mm,
  toptitle=0.8mm,
  bottomtitle=0.8mm,
  pad at break*=1mm
]

\small
\sloppy
\setlength{\parindent}{0pt}
\setlength{\parskip}{3pt}

\textbf{Video-Only Diagnostic Evaluation Prompt}

\medskip

\textbf{\#\#\# Role and Task}

\textit{You are a professional QA auditor responsible for diagnosing failures in
AI-generated interactive web applications. For every supplied single-choice diagnostic question, compare all candidate
failure modes and select exactly one option that is most directly supported by
the available video evidence and most relevant to the intended user workflow.
Answer every question exactly once.}

\medskip

\textbf{\#\#\# 1. Input}

You will receive:

\begin{enumerate}
  \setlength{\itemsep}{1pt}
  \setlength{\parskip}{0pt}
  \setlength{\parsep}{0pt}
  \setlength{\topsep}{2pt}

  \item a screen-recording video showing a human operating the generated web
  application; and

  \item a set of diagnostic questions containing the question ID, target
  evaluation dimension, intended user workflow, question text, and candidate
  failure modes with definitions.
\end{enumerate}

\medskip

\textbf{\#\#\# 2. Evidence Priority}

\textbf{Primary Evidence: Operation Video.}

Use the video as the sole source for judging rendered appearance, layout,
readability, motion, user actions, interaction responses, state changes,
navigation, and workflow progression.

Base all judgments on what is visibly rendered and experienced by the user in
the video.

\medskip

\textbf{\#\#\# 3. Evaluation Rules}

\begin{enumerate}
  \setlength{\itemsep}{3pt}
  \setlength{\parskip}{0pt}
  \setlength{\parsep}{0pt}
  \setlength{\topsep}{2pt}

  \item \textbf{Complete and dimension-specific evaluation.}
  Evaluate every question independently according to its target dimension,
  intended workflow, and candidate definitions. Do not consider unrelated
  defects.

  \item \textbf{Internal candidate comparison.}
  Inspect every candidate before selecting an answer, but do not include the
  option-by-option comparison in the final output.

  \item \textbf{Direct evidence only.}
  Select an option only when it is supported by observable evidence in the
  video. Do not rely on speculation or assumed behavior.

  \item \textbf{No unsupported interaction assumptions.}
  Do not infer the result of an action that is not performed or visibly
  demonstrated in the video.

  \item \textbf{Specific evidence.}
  Support the selected option with a concise description of the observed
  layout, interaction, animation, navigation, state, content, or workflow
  failure.

  \item \textbf{Single-choice decision.}
  Select exactly one supplied \texttt{option\_id}. If several options appear
  related, choose the one with the clearest evidence, most precise definition
  match, and greatest relevance to the intended workflow. Briefly acknowledge
  genuine ambiguity in the reasoning.

  \item \textbf{Excluded considerations.}
  Do not evaluate recording quality, video compression, capture artifacts,
  operator skill, failures outside the target dimension, or failure types not
  included among the candidates.
\end{enumerate}

\medskip

\textbf{\#\#\# 4. Diagnostic Question Format}

Each question follows this structure:

\medskip

\textbf{Question ID:} \texttt{<QUESTION\_ID>}

\textbf{Question Type:} \texttt{single\_choice}

\textbf{Target Evaluation Dimension:}\\
\texttt{<TARGET\_EVALUATION\_DIMENSION>}

\textbf{Generation Intent (User Workflow):}\\
\texttt{<GENERATION\_INTENT\_USER\_WORKFLOW>}

\textbf{Question:}\\
\texttt{<QUESTION\_TEXT>}

\textbf{Candidate Options:}

\begin{itemize}
  \setlength{\itemsep}{2pt}
  \setlength{\parskip}{0pt}
  \setlength{\parsep}{0pt}
  \setlength{\topsep}{2pt}

  \item \textbf{A:} \texttt{<OPTION\_A\_FAILURE\_MODE>}\\
  \textbf{Definition:} \texttt{<OPTION\_A\_DEFINITION>}

  \item \textbf{B:} \texttt{<OPTION\_B\_FAILURE\_MODE>}\\
  \textbf{Definition:} \texttt{<OPTION\_B\_DEFINITION>}

  \item \textbf{C:} \texttt{<OPTION\_C\_FAILURE\_MODE>}\\
  \textbf{Definition:} \texttt{<OPTION\_C\_DEFINITION>}

  \item \textbf{D:} \texttt{<OPTION\_D\_FAILURE\_MODE>}\\
  \textbf{Definition:} \texttt{<OPTION\_D\_DEFINITION>}
\end{itemize}

\medskip

\textbf{\#\#\# 5. Output Requirements}

Return only one valid JSON object in the following form:

\medskip

{\ttfamily\footnotesize
\{\\
\hspace*{1em}"answers": [\\
\hspace*{2em}\{\\
\hspace*{3em}"question\_id": "<QUESTION\_ID>",\\
\hspace*{3em}"reasoning":
"<CONCISE\_EVIDENCE\_BASED\_REASONING>",\\
\hspace*{3em}"selected\_option\_ids": ["A"],\\
\hspace*{3em}"evidence": \{\\
\hspace*{4em}"option\_id": "A",\\
\hspace*{4em}"video\_description":
"<BRIEF\_DESCRIPTION\_OF\_THE\_OBSERVED\_EVIDENCE>"\\
\hspace*{3em}\}\\
\hspace*{2em}\}\\
\hspace*{1em}]\\
\}
}

\medskip

The \texttt{answers} array must contain exactly one object for every supplied
\texttt{question\_id}. Each object must contain only
\texttt{question\_id}, \texttt{reasoning},
\texttt{selected\_option\_ids}, and \texttt{evidence}, in that order.

The \texttt{selected\_option\_ids} field must contain exactly one supplied
option ID. The \texttt{evidence} object must contain only
\texttt{option\_id} and \texttt{video\_description}.

Do not output unselected options, internal candidate comparisons, additional
fields, summaries, Markdown code fences, or text outside the JSON object.

\medskip

\textbf{---TASK:}

Evaluate every supplied diagnostic question and return the required JSON
object now.

\end{tcolorbox}
\end{center}

\refstepcounter{figure}
\label{fig:video_only_diagnostic_prompt}

{\small
\centering
Figure~\thefigure:
\textbf{Prompt used for video-only diagnostic evaluation}
\par
}

\par\medskip

\begin{center}
\begin{tcolorbox}[
  enhanced jigsaw,
  breakable,
  width=0.88\textwidth,
  colback=blue!2,
  colframe=blue!55!black,
  colbacktitle=blue!55!black,
  coltitle=white,
  title=\textbf{System Instruction},
  fonttitle=\small\bfseries,
  arc=1mm,
  boxrule=0.6pt,
  left=2mm,
  right=2mm,
  top=1.5mm,
  bottom=1.5mm,
  toptitle=0.8mm,
  bottomtitle=0.8mm,
  pad at break*=1mm
]

\small
\sloppy
\setlength{\parindent}{0pt}
\setlength{\parskip}{3pt}

\textbf{HTML-Only Diagnostic Evaluation Prompt}

\medskip

\textbf{\#\#\# Role and Task}

\textit{You are a professional QA auditor responsible for diagnosing failures in
AI-generated interactive web applications from their HTML/CSS/JavaScript
source code. For every supplied single-choice diagnostic question, compare all candidate
failure modes and select exactly one option that is most directly supported by
the available source-code evidence and most relevant to the intended user
workflow. Answer every question exactly once.}

\medskip

\textbf{\#\#\# 1. Input}

You will receive:

\begin{enumerate}
  \setlength{\itemsep}{1pt}
  \setlength{\parskip}{0pt}
  \setlength{\parsep}{0pt}
  \setlength{\topsep}{2pt}

  \item the complete HTML/CSS/JavaScript source code of the generated web
  application; and

  \item a set of diagnostic questions containing the question ID, target
  evaluation dimension, intended user workflow, question text, and candidate
  failure modes with definitions.
\end{enumerate}

\medskip

\textbf{\#\#\# 2. Evidence Source}

\textbf{Only Evidence: HTML/CSS/JavaScript.}

Use the complete source code as the only evidence for judging DOM structure,
CSS properties, displayed text, asset references, event handlers, state
variables, state-update logic, validation rules, navigation targets, and
implemented client-side behavior.

No operation video, screenshot, rendered webpage, browser runtime, network
log, or interaction trace is available.

The diagnostic question may mention a video, visual observation, user action,
or timestamp because the same questions are used across input modalities.
Interpret the underlying failure-diagnosis question using only the supplied
source code. Do not claim that any visual or runtime behavior was directly
observed.

Static source code may not conclusively establish browser-dependent,
server-dependent, network-dependent, timing-dependent, or
user-action-dependent behavior.

\medskip

\textbf{\#\#\# 3. Evaluation Rules}

\begin{enumerate}
  \setlength{\itemsep}{3pt}
  \setlength{\parskip}{0pt}
  \setlength{\parsep}{0pt}
  \setlength{\topsep}{2pt}

  \item \textbf{Complete and dimension-specific evaluation.}
  Evaluate every question independently according to its target dimension,
  intended workflow, and candidate definitions. Do not consider unrelated
  defects.

  \item \textbf{Internal candidate comparison.}
  Inspect every candidate before selecting an answer, but do not include the
  option-by-option comparison in the final output.

  \item \textbf{Direct source-code evidence only.}
  Select an option based on concrete evidence explicitly supported by the
  source code. When possible, identify the relevant element, selector, CSS
  property, text content, asset path, event handler, state update, validation
  condition, or navigation target.

  \item \textbf{No runtime fabrication.}
  Do not invent rendered appearance, user actions, interaction results,
  animation behavior, navigation outcomes, network responses, or timestamps.
  Do not describe a possible implementation risk as a directly observed
  runtime failure.

  \item \textbf{Specific evidence.}
  Support the selected option with a concise description of the relevant
  HTML/CSS/JavaScript implementation evidence.

  \item \textbf{Single-choice decision.}
  Select exactly one supplied \texttt{option\_id}. If several options appear
  related, choose the one with the strongest explicit source-code evidence,
  most precise definition match, and greatest relevance to the intended user
  workflow.

  If no option is conclusively established by static source code, select the
  option with the strongest available source-code support. This selection
  represents relative evidential support rather than confirmation of an
  observed runtime failure.

  \item \textbf{Excluded considerations.}
  Do not evaluate unavailable visual or runtime evidence, failures outside the
  target dimension, or failure types not included among the candidates.
\end{enumerate}

\medskip

\textbf{\#\#\# 4. Diagnostic Question Format}

Each question follows this structure:

\medskip

\textbf{Question ID:} \texttt{<QUESTION\_ID>}

\textbf{Question Type:} \texttt{single\_choice}

\textbf{Target Evaluation Dimension:}\\
\texttt{<TARGET\_EVALUATION\_DIMENSION>}

\textbf{Generation Intent (User Workflow):}\\
\texttt{<GENERATION\_INTENT\_USER\_WORKFLOW>}

\textbf{Question:}\\
\texttt{<QUESTION\_TEXT>}

\textbf{Candidate Options:}

\begin{itemize}
  \setlength{\itemsep}{2pt}
  \setlength{\parskip}{0pt}
  \setlength{\parsep}{0pt}
  \setlength{\topsep}{2pt}

  \item \textbf{A:} \texttt{<OPTION\_A\_FAILURE\_MODE>}\\
  \textbf{Definition:} \texttt{<OPTION\_A\_DEFINITION>}

  \item \textbf{B:} \texttt{<OPTION\_B\_FAILURE\_MODE>}\\
  \textbf{Definition:} \texttt{<OPTION\_B\_DEFINITION>}

  \item \textbf{C:} \texttt{<OPTION\_C\_FAILURE\_MODE>}\\
  \textbf{Definition:} \texttt{<OPTION\_C\_DEFINITION>}

  \item \textbf{D:} \texttt{<OPTION\_D\_FAILURE\_MODE>}\\
  \textbf{Definition:} \texttt{<OPTION\_D\_DEFINITION>}
\end{itemize}

\medskip

\textbf{\#\#\# 5. Output Requirements}

Return only one valid JSON object in the following form:

\medskip

{\ttfamily\footnotesize
\{\\
\hspace*{1em}"answers": [\\
\hspace*{2em}\{\\
\hspace*{3em}"question\_id": "<QUESTION\_ID>",\\
\hspace*{3em}"reasoning":
"<CONCISE\_EVIDENCE\_BASED\_REASONING>",\\
\hspace*{3em}"selected\_option\_ids": ["A"],\\
\hspace*{3em}"evidence": \{\\
\hspace*{4em}"option\_id": "A",\\
\hspace*{4em}"video\_description":
"<BRIEF\_DESCRIPTION\_OF\_THE\_SOURCE\_CODE\_EVIDENCE>"\\
\hspace*{3em}\}\\
\hspace*{2em}\}\\
\hspace*{1em}]\\
\}
}

\medskip

The \texttt{answers} array must contain exactly one object for every supplied
\texttt{question\_id}. Each object must contain only
\texttt{question\_id}, \texttt{reasoning},
\texttt{selected\_option\_ids}, and \texttt{evidence}, in that order.

The \texttt{reasoning} field must explain why the selected option has the
strongest explicit HTML/CSS/JavaScript support without claiming that the
failure was visually or operationally observed.

The \texttt{selected\_option\_ids} field must contain exactly one supplied
option ID. The \texttt{evidence} object must contain only
\texttt{option\_id} and \texttt{video\_description}.

The \texttt{video\_description} field is retained for output compatibility but
must contain a concise description of the source-code evidence rather than
video evidence.

Do not output unselected options, internal candidate comparisons, additional
fields, summaries, Markdown code fences, or text outside the JSON object.

\medskip

\textbf{---TASK:}

Evaluate every supplied diagnostic question and return the required JSON
object now.

\end{tcolorbox}
\end{center}

\refstepcounter{figure}
\label{fig:html_only_diagnostic_prompt}

{\small
\centering
Figure~\thefigure:
\textbf{Prompt used for HTML-only diagnostic evaluation}
\par
}

\newtcolorbox{v2lensprompt}{
  enhanced jigsaw,
  breakable,
  width=0.88\textwidth,
  colback=blue!2,
  colframe=blue!55!black,
  colbacktitle=blue!55!black,
  coltitle=white,
  title=\textbf{System Instruction},
  fonttitle=\small\bfseries,
  arc=1mm,
  boxrule=0.6pt,
  left=2mm,
  right=2mm,
  top=1.5mm,
  bottom=1.5mm,
  toptitle=0.8mm,
  bottomtitle=0.8mm,
  pad at break*=1mm
}

\par\medskip

\begin{center}
\begin{v2lensprompt}

\small
\sloppy
\setlength{\parindent}{0pt}
\setlength{\parskip}{3pt}

\textbf{V2Lens Video Analyst Prompt}

\medskip

\textbf{\#\#\# Role and Task}

\textit{You are the \textbf{Video Analyst} in V2Lens. Examine the complete interaction
video and answer each single-choice diagnostic question.}

For every question:

\begin{itemize}
  \setlength{\itemsep}{1pt}
  \setlength{\parskip}{0pt}
  \setlength{\topsep}{2pt}

  \item determine whether each candidate failure mode is
  \texttt{present} or \texttt{absent};
  \item provide concise video evidence for each option; and
  \item select the single option that most directly explains the observed
  failure.
\end{itemize}

Several failure modes may be visible, but the initial answer must be the one
most relevant to the intended user workflow.

\medskip

\textbf{\#\#\# 1. Evidence Rules}

Use only directly observable user actions, interface responses, state changes,
navigation behavior, and workflow outcomes from the video.

Ignore recording quality, compression artifacts, operator skill, and
recording-tool overlays. Do not infer behavior from interactions that are not
shown.

\medskip

\textbf{\#\#\# 2. Input}

\textbf{Video ID:} \texttt{<VIDEO\_ID>}

\textbf{Question and Intended Workflow:}\\
\texttt{<QUESTION\_AND\_WORKFLOW>}

\textbf{Candidate Options and Definitions:}\\
\texttt{<CANDIDATE\_OPTIONS\_AND\_DEFINITIONS>}

\medskip

\textbf{\#\#\# 3. Output Requirements}

Return only one valid JSON object:

\medskip

{\ttfamily\footnotesize
\{\\
\hspace*{1em}"question\_id": "<QUESTION\_ID>",\\
\hspace*{1em}"initial\_answer": "A",\\
\hspace*{1em}"reasoning": "<CONCISE\_VIDEO\_BASED\_REASONING>",\\
\hspace*{1em}"option\_assessments": [\\
\hspace*{2em}\{\\
\hspace*{3em}"option\_id": "A",\\
\hspace*{3em}"verdict": "present|absent",\\
\hspace*{3em}"video\_evidence": "<OBSERVED\_INTERACTION\_EVIDENCE>"\\
\hspace*{2em}\}\\
\hspace*{1em}]\\
\}
}

Include one assessment for every candidate option. Do not output Markdown or
additional text.

\medskip

\textbf{---TASK:}

Analyze the video and return the required JSON object now.

\end{v2lensprompt}
\end{center}

\refstepcounter{figure}
\label{fig:v2lens_video_analyst_prompt}

{\small
\centering
Figure~\thefigure:
\textbf{Prompt used by the V2Lens Video Analyst}
\par
}

\par\medskip

\begin{center}
\begin{v2lensprompt}

\small
\sloppy
\setlength{\parindent}{0pt}
\setlength{\parskip}{3pt}

\textbf{V2Lens HTML Verifier Prompt}

\medskip

\textbf{\#\#\# Role and Task}

\textit{You are the \textbf{HTML Verifier} in V2Lens. Inspect the webpage's
HTML/CSS/JavaScript source and evaluate the implementation evidence for every
candidate failure mode.}

For each option, determine whether the source code:

\begin{itemize}
  \setlength{\itemsep}{1pt}
  \setlength{\parskip}{0pt}
  \setlength{\topsep}{2pt}

  \item \texttt{supports} the diagnosis;
  \item \texttt{contradicts} the diagnosis; or
  \item is \texttt{inconclusive}.
\end{itemize}

\medskip

\textbf{\#\#\# 1. Evidence Rules}

Inspect relevant DOM structure, CSS rules, text, assets, event handlers,
validation conditions, navigation targets, and state-update logic.

Use only explicit source-code evidence. Code findings provide implementation
context but do not independently prove that a failure appeared during the
recorded interaction.

\medskip

\textbf{\#\#\# 2. Input}

\textbf{Question and Intended Workflow:}\\
\texttt{<QUESTION\_AND\_WORKFLOW>}

\textbf{Candidate Options and Definitions:}\\
\texttt{<CANDIDATE\_OPTIONS\_AND\_DEFINITIONS>}

\textbf{Video Analyst Assessments:}\\
\texttt{<VIDEO\_ANALYST\_ASSESSMENTS>}

\textbf{HTML/CSS/JavaScript Source:}\\
\texttt{<WEBPAGE\_SOURCE>}

\medskip

\textbf{\#\#\# 3. Output Requirements}

Return only one valid JSON object:

\medskip

{\ttfamily\footnotesize
\{\\
\hspace*{1em}"question\_id": "<QUESTION\_ID>",\\
\hspace*{1em}"findings": [\\
\hspace*{2em}\{\\
\hspace*{3em}"option\_id": "A",\\
\hspace*{3em}"code\_verdict":
"supports|contradicts|inconclusive",\\
\hspace*{3em}"code\_evidence": "<RELEVANT\_IMPLEMENTATION\_EVIDENCE>"\\
\hspace*{2em}\}\\
\hspace*{1em}]\\
\}
}

Include one finding for every candidate option. Do not output Markdown or
additional text.

\medskip

\textbf{---TASK:}

Inspect the source code and return the required JSON object now.

\end{v2lensprompt}
\end{center}

\refstepcounter{figure}
\label{fig:v2lens_html_verifier_prompt}

{\small
\centering
Figure~\thefigure:
\textbf{Prompt used by the V2Lens HTML Verifier}
\par
}

\par\medskip

\begin{center}
\begin{v2lensprompt}

\small
\sloppy
\setlength{\parindent}{0pt}
\setlength{\parskip}{3pt}

\textbf{V2Lens Evidence Judge Prompt}

\medskip

\textbf{\#\#\# Role and Task}

\textit{You are the \textbf{Evidence Judge} in V2Lens. Integrate Video Analyst assessments, HTML Verifier findings, and available video and source-code evidence.
Retain the initial answer unless another candidate is more strongly supported
by the collected evidence. When a stronger alternative exists, produce a
challenge answer and the evidence supporting it.}

\medskip

\textbf{\#\#\# 1. Evidence Rules}

The interaction video is the primary evidence, while source code provides
supporting implementation evidence.

When necessary, use frame extraction, crop and zoom, visual measurement, or
HTML inspection to resolve insufficient or conflicting evidence.

A challenge must be supported by direct observable behavior and must not rely
only on indirect source-code clues.

\medskip

\textbf{\#\#\# 2. Input}

\textbf{Question, Workflow, and Candidate Options:}\\
\texttt{<QUESTION\_WORKFLOW\_AND\_OPTIONS>}

\textbf{Initial Answer:}\\
\texttt{<VIDEO\_ANALYST\_INITIAL\_ANSWER>}

\textbf{Video Analyst Assessments:}\\
\texttt{<VIDEO\_ANALYST\_ASSESSMENTS>}

\textbf{HTML Verifier Findings:}\\
\texttt{<HTML\_VERIFIER\_FINDINGS>}

\textbf{Interaction Video and Webpage Source:}\\
\texttt{<VIDEO\_AND\_SOURCE\_CONTEXT>}

\medskip

\textbf{\#\#\# 3. Output Requirements}

Return only one valid JSON object:

\medskip

{\ttfamily\footnotesize
\{\\
\hspace*{1em}"question\_id": "<QUESTION\_ID>",\\
\hspace*{1em}"decision": "retain|challenge",\\
\hspace*{1em}"challenge\_answer": "<OPTION\_ID\_OR\_EMPTY>",\\
\hspace*{1em}"reasoning": "<CONCISE\_CROSS\_MODAL\_REASONING>",\\
\hspace*{1em}"supporting\_evidence": "<DIRECT\_EVIDENCE\_FOR\_THE\_DECISION>"\\
\}
}

Use \texttt{retain} and an empty \texttt{challenge\_answer} when no alternative
is more strongly supported. Do not output Markdown or additional text.

\medskip

\textbf{---TASK:}

Evaluate the collected evidence and return the required JSON object now.

\end{v2lensprompt}
\end{center}

\refstepcounter{figure}
\label{fig:v2lens_evidence_judge_prompt}

{\small
\centering
Figure~\thefigure:
\textbf{Prompt used by the V2Lens Evidence Judge}
\par
}

\par\medskip

\begin{center}
\begin{v2lensprompt}

\small
\sloppy
\setlength{\parindent}{0pt}
\setlength{\parskip}{3pt}

\textbf{V2Lens Reviewer Prompt}

\medskip

\textbf{\#\#\# Role and Task}

\textit{You are the independent \textbf{Reviewer} in V2Lens. Compare two anonymized
candidate answers using the original interaction video, webpage source,
diagnostic question, and candidate definitions.
You are not told which candidate is the initial answer or the challenge
answer. You cannot access the Evidence Judge's reasoning or supporting
evidence.
Apply the same standard to both candidates and determine which is more
directly supported.}

\medskip

\textbf{\#\#\# 1. Evidence Rules}

Use the interaction video as the primary evidence and source code as supporting
evidence. Inspect additional frames, visual details, measurements, or source
code when necessary.

Do not prefer a candidate because it was routed for review. Judge both
independently.

\medskip

\textbf{\#\#\# 2. Input}

\textbf{Question, Workflow, and Candidate Definitions:}\\
\texttt{<QUESTION\_WORKFLOW\_AND\_DEFINITIONS>}

\textbf{Candidate 1:}\\
\texttt{<ANONYMIZED\_CANDIDATE\_1>}

\textbf{Candidate 2:}\\
\texttt{<ANONYMIZED\_CANDIDATE\_2>}

\textbf{Interaction Video and Webpage Source:}\\
\texttt{<VIDEO\_AND\_SOURCE\_CONTEXT>}

\medskip

\textbf{\#\#\# 3. Output Requirements}

Return only one valid JSON object:

\medskip

{\ttfamily\footnotesize
\{\\
\hspace*{1em}"question\_id": "<QUESTION\_ID>",\\
\hspace*{1em}"preferred\_candidate":
"candidate\_1|candidate\_2|undetermined",\\
\hspace*{1em}"reasoning": "<CONCISE\_INDEPENDENT\_EVIDENCE\_BASED\_REASONING>"\\
\}
}

Use \texttt{undetermined} when the evidence does not reliably distinguish the
two candidates. Do not output Markdown or additional text.

\medskip

\textbf{---TASK:}

Independently review the two candidates and return the required JSON object
now.

\end{v2lensprompt}
\end{center}

\refstepcounter{figure}
\label{fig:v2lens_reviewer_prompt}

{\small
\centering
Figure~\thefigure:
\textbf{Prompt used by the V2Lens Reviewer}
\par
}

\par\medskip